\pdfoutput=1

\documentclass[11pt]{article}

\usepackage[]{ACL2023}
\usepackage{tikz}
\usepackage{amsmath}
\usepackage{amsfonts}
\usepackage{arydshln}
\usepackage{times}
\usepackage{latexsym}
\usepackage{graphicx}
\usepackage{subcaption}
\usepackage[T1]{fontenc}

\usepackage[utf8]{inputenc}
\usepackage{caption}
\usepackage{subcaption}
\usepackage{microtype}
\usepackage{inconsolata}
\usepackage{enumitem}

\usepackage{booktabs}

\title{Text Embedding Inversion Security for Multilingual Language Models} 

\author{Yiyi Chen \hfill Heather Lent \hfill Johannes Bjerva \\
          Department of Computer Science, Aalborg University, Denmark \\
           \texttt{\{yiyic, hcle, jbjerva\}@cs.aau.dk} \\
           }

\begin{document}
\maketitle
\begin{abstract}

Textual data is often represented as real-numbered embeddings in NLP, particularly with the popularity of large language models (LLMs) and Embeddings as a Service (EaaS). 
However, storing sensitive information as embeddings can be susceptible to security breaches, as research shows that text can be reconstructed from embeddings, even without knowledge of the underlying model. 
While defence mechanisms have been explored, these are exclusively focused on English, leaving other languages potentially exposed to attacks.
This work explores LLM security through \textit{multilingual} embedding inversion. 
We define the problem of black-box multilingual and cross-lingual inversion attacks, and explore their potential implications.
Our findings suggest that multilingual LLMs may be more vulnerable to inversion attacks, in part because English-based defences may be ineffective.
To alleviate this, we propose a simple masking defense effective for both monolingual and multilingual models.
This study is the first to investigate multilingual inversion attacks, shedding light on the differences in attacks and defenses across monolingual and multilingual settings.

\end{abstract}

\section{Introduction}

\begin{figure*}[!t]
    \centering
    \includegraphics[width=\linewidth]{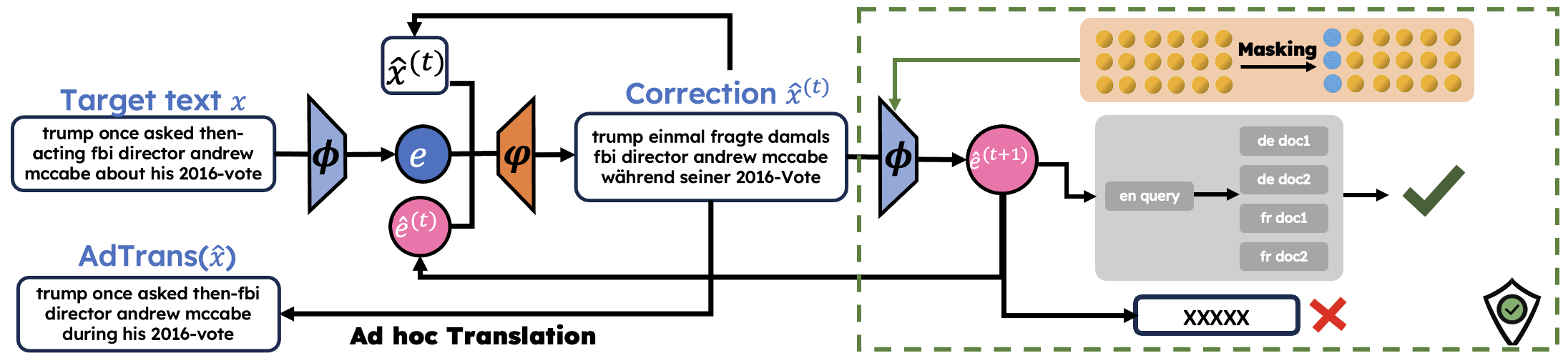}
    \caption{Overview of Multilingual Vec2Text, extending Vec2Text~\citep{morris-etal-2023-text} with Ad hoc Translation and Masking Defense Mechanism (outlined in the green dashed line frame).
    Given access to a target embedding \textcolor{blue}{$e$} and query access to the embedder \textcolor{blue}{$\mathbf{\phi}$} via an EaaS API, the inversion model \textcolor{orange}{\textbf{$\psi$}} iteratively generates hypotheses \textcolor{magenta}{\textbf{$\hat{e}$}} to attain the target. 
    The generated text $\hat{x}$ is in German, and translated to English (AdTrans($\hat{x}$)), to be compared with the target text $x$. 
    The masking defense serves as an effective defense against inversion attacks while preserving utility in NLP tasks such as retrieval. }
    \label{fig:overview}
\end{figure*}

Industrial applications of natural language processing (NLP) typically utilize language models (LMs) and often rely on vector databases via frameworks such as Embeddings as a Service (EaaS).
In this context, sentence embeddings are stored in a remote database, as opposed to raw text, allowing end-users to efficiently search across condensed representations.
As embeddings are not human-readable, security of the encoded information may be naively assumed, however 
recent works have demonstrated that embeddings are no safer than raw text; they are susceptible to \textit{inversion attacks}, whereby a malicious actor can train models to decode embeddings, thus exposing private information \cite{10.1145/3372297.3417270, morris-etal-2023-text, zhou-etal-2023-textobfuscator}. Concretely, after gaining access to embeddings and the black-box embedder via the EaaS API, the malicious actor can train an external model, which approximates the inversion function that reconstructs the text from the embeddings. As such, there is a substantial threat to privacy if malicious actors are able to eavesdrop on communication channels between EaaS providers and customers, as illustrated in Figure~\ref{fig:overview}.

\begin{figure}[!t]
    \centering
    \includegraphics[width=\linewidth]{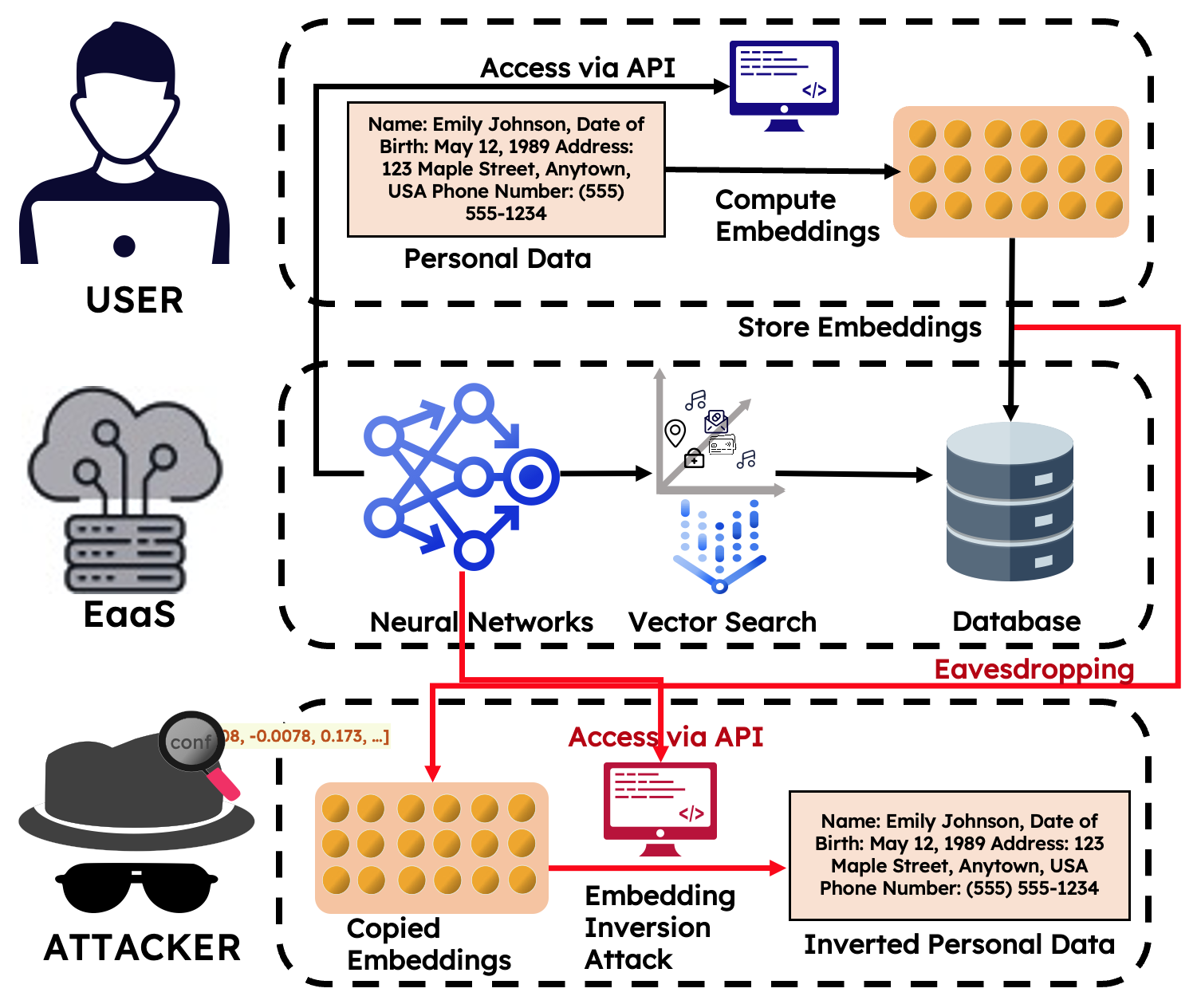}
    \caption{Schematic overview of a text embedding inversion attack. A user accesses an EaaS provider, while an attacker is eavesdropping. Although the attacker has no direct access to the embedding model, they can reliably decode the information stored in the embeddings.}
    \label{fig:attack_secenario}
\end{figure}

Previous work has shown that an exact match for data recreation can be obtained in specific settings, albeit with the limitation of assuming monolingual English models and embeddings \citep{morris-etal-2023-text}.
However, in real-world scenarios, eavesdroppers may not know the source language of the encoded text, as EaaS providers can have international clientele. 
Thus to assess the current level of risk posed to multilingual LMs, we introduce  \textit{multilingual} inversion attacks.
As the first ever study in this direction, we focus specifically on exact text reconstruction, assuming that the language of a target embedding is unknown. 
Leveraging a state-of-the-art multilingual black-box encoder, we find that the trained model can reconstruct texts in certain languages more effectively than monolingual counterparts.
Additionally, we also introduce \textit{cross-lingual} inversion attacks, 
to ascertain whether inversion attacks can be successful when the target language is unknown by the attacker.
We thus attempt cross-lingual text reconstruction 
(i.e., reconstructing German text with a model not trained on German reconstruction),
introducing an \textit{Ad hoc Translation} method to overcome the evaluation limitation of current string-matching metrics in this cross-lingual scenario. 
Finally, we assess the efficacy of an existing defense method by \citet{morris-etal-2023-text}, ultimately finding that defenses intended for monolingual models fall short in protecting multilingual models. 
To this end, we introduce simple \textit{masking defense}, which proves effective for both monolingual and multilingual models, and which also does not require additional model training.
All our trained inversion models~\footnote{\url{https://huggingface.co/yiyic/}}
and code
\footnote{\url{https://github.com/siebeniris/MultiVec2Text/}} are open source, encouraging the research community to engage in development of defenses for vulnerable multilingual models.

\section{Related Work} 
Models are well known to memorize training data, and are therefore susceptible to leaking private information \cite{DBLP:journals/corr/ShokriSS16,DBLP:journals/corr/abs-1802-08232,Nasr_2019}. 
As such, there is increased research interest in exploring this vulnerability to \textit{inversion attacks} from the perspective of cyber-security, simulating attacks against models to recreate sensitive training data. 
Work in this direction has been conducted across various domains of machine learning, such as computational genetics \cite{10.5555/2671225.2671227}, computer vision \cite{10.1145/2810103.2813677}, and more recently NLP \cite{10.1145/3372297.3417270}. 
Generally, such works at the intersection of machine learning and cyber-security (e.g., on inversion attacks or adversarial attacks) make assumptions about the imagined attacker's levels of access to the victim model. 
White-box scenarios assume attacker access to the full model \cite{wallace-etal-2019-universal, tsymboi-etal-2023-layerwise}, resulting in many possible attack surfaces.
Previous works in NLP have shown that it is possible to retrieve sensitive training data by attacking models directly \cite{10.5555/2671225.2671227, 10.1145/2810103.2813677}, attacking gradients \cite{NEURIPS2019_60a6c400, deng-etal-2021-tag-gradient}, as well as through leveraging leaked hidden states \cite{li-etal-2022-dont}. 
Meanwhile, black-box attacks assume an attacker has no knowledge of the underlying model itself, and can only interact with models at the most abstracted level (e.g., provide input and register output through an API). For example, \citet{DBLP:journals/corr/abs-2012-07805} are able to extract sensitive training data (e.g., names and phone numbers) from GPT-2 \cite{Radford2019LanguageMA}, by first generating data from the model and then using membership inference attacks to filter utterances likely to be part of the original training data. 

In embedding inversion attacks, an imagined attacker aims to recreate text from the distributed representations. 
As opposed to a machine translation setting, this scenario assumes no access to a source text $x$ to condition on, and the goal is not to decode a translation of $x$, but rather to recreate the exact text of $x$ --- with no input other than the embedding $\phi(x)$, given $\phi$ as an encoder.
\citet{10.1145/3372297.3417270} showed that 50\%--70\% percent of tokens could be recovered in such a setting. 
Subsequent attacks have further improved over this metric, with newer approaches now able to retrieve entire sentences of encoded text \cite{DBLP:journals/corr/abs-2109-10104, hayet-etal-2022-invernet, morris-etal-2023-text, li-etal-2023-sentence}. 
Existing defense mechanisms include randomly perturbing embeddings \cite{zhou-etal-2023-textobfuscator} and parameter-efficient fine-tuning \cite{zhang-etal-2023-fedpetuning}. 
Other methods for securing embeddings include encryption \cite{huang-etal-2020-texthide, xie-hong-2021-reconstruction} and differential privacy \cite{Lyu2020DifferentiallyPR}. 
However, until embedding privacy is ensured, inversion attacks will remain a threat, necessitating further investigation.

Finally, previous works on embedding inversion have been confined to monolingual settings concerning English~\citep{10.1145/3372297.3417270,Lyu2020DifferentiallyPR, hayet-etal-2022-invernet, parikh-etal-2022-canary, kim-etal-2022-toward, morris-etal-2023-text, zhou-etal-2023-textobfuscator, li-etal-2023-sentence}.
This leaves defenses for non-English languages and multilingual models unexplored, potentially compromising model security for those languages.
As a result, the vulnerability of multilingual models and non-English models remains an open question.

\section{Methodology}\label{sec:methodology}

In this work, we consider a scenario where a malicious actor has illegitimately obtained both embeddings and API access to the black-box encoder, as shown in Figure~\ref{fig:attack_secenario}.
To gauge the vulnerability of multilingual models against black-box embedding inversion attacks, we build upon previous work by \citet{morris-etal-2023-text}, extending their attack method to a multilingual setting, aiming to invert sentence embeddings produced by a multilingual model. 
We define the attack scenario formally as follows: given a sensitive text sequence $x$ and a black-box encoder $\phi$, the goal is to recover $x$ from the embedding obtained via $\phi(x)$ using an external attacker model $\psi$. However, we can only access $\phi$ through an EaaS API, and its architecture and parameters are inaccessible.
To this end, we explore the efficacy of existing defenses in this scenario, and introduce a novel defense mechanism.

We approach embedding inversion attacks in the context of text generation, considering the generation models' efficacy in such attacks~\citep{li-etal-2023-sentence,morris-etal-2023-text}. 
In this scenario, the generation model $\psi$ conditions what information can be encoded and decoded, with consequences for text reconstruction.
For example, if $\psi$ is solely pre-trained on Latin script, it cannot handle Cyrillic or Devanagari scripts. 
Consequently, reconstructing text in unknown scripts is presently infeasible, and whether text in unknown scripts can be reconstructed remains unexplored.
Hence, our study investigates text reconstruction in unknown \textit{languages} within the same script (i.e., Latin).

\paragraph{Multilingual Inversion Attacks} 
Compared to monolingual embedding inversion, investigating multilingual inversion attacks introduces significant complexity, as each language space of $\psi$, $\phi$, $x$, and training data is crucial. 
For instance, the training scale for attacker models increases with the number of languages and controlled parameters, such as maximal sequence length (cf. Section~\ref{sec:exp}).

We explore the potential of multilingual embedding inversion assuming unlimited queries can be sent to the black-box $\phi$, obtaining embeddings $\phi(x)$ for $x\in \mathbb{D}$, where $\mathbb{D}$ is the training dataset. Following the approximation approach from \citet{morris-etal-2023-text}, we search for text $\hat{x}$ closest to the target embedding $e$ under $\phi$ using the formula:

\begin{equation}\label{eq:cos_equation}
    \hat{x} = arg\max_{x}\cos(\phi(x), e)
\end{equation}

In particular, as illustrated in Figure~\ref{fig:overview}, the training and inference of the inversion model are conditioned on the previous output. At correction step $t+1$, the model takes the concatenation of the previous output $\hat{x}^{(t)}$, hypothesis embedding $\hat{e}^{(t)}$, and target embedding $e$. With this context noted, the multilingual embedding inversion attack is composed of the following steps: 

\begin{itemize}[noitemsep,topsep=1pt]
    \item \textbf{Base model Model Training:} Develop an attacker model $\psi$ based on a text generation model pre-trained on the same language scripts;
    \item \textbf{Correction Model Training:} Train $\psi$ by querying the black-box embedding model $\phi$ with text $x\in \mathbb{D}$, resulting in $\hat{x}$ optimized using Eq.~\ref{eq:cos_equation} (correction step 1).
    \item \textbf{Inference:} Execute embedding inversion attacks on texts in the target language $l_t$ using the trained inversion model $\psi$. Further optimization (correction steps $>1$) is performed with Eq.\ref{eq:cos_equation}, combined with beam search at sequence level.
\end{itemize}

\paragraph{Cross-Lingual Inversion Attacks} 
In a multilingual setting we assume that the inversion model is trained on several languages, including the target text language $l_{t}$.
However, this is an unrealistic setting which requires immense computational resources.
We therefore investigate a cross-lingual setting, in which  the aggressor does not know the true language of the target text $l_{t}$. 
Concretely, we investigate the extent it is possible to execute inversion attacks leveraging a monolingual inversion model trained on a \textit{different} source language $l_s$ than the target $l_{t}$, thus introducing a cross-lingual attack. 
As the text generated by the monolingual inversion model will be in $l_s$, current string-matching metrics for evaluating inversion attacks, such as BLEU, are not applicable here, as there will be little or no overlap between the $l_s$ and $l_{t}$ strings, even when the underlying meaning of the two is the same. 
In order to evaluate the success of the cross-lingual inversion model, we propose a post-intervention strategy \textbf{Ad hoc Translation} (AdTrans), as shown in Figure~\ref{fig:overview}. In this setup, the generated text is first translated from $l_s$ in $l_t$ using EasyNMT~\footnote{\url{https://github.com/UKPLab/EasyNMT}}. Then the translated text is evaluated against the target text, to verify whether the inverted text in $l_s$ can indeed uncover the target text in unknown $l_{t}$ (cf. Section~\ref{sec:cross-lingual}). As AdTrans hinges upon the availability of a reliable machine translation model for the pertinent languages, this use case highlights the existing limitations in current evaluation metrics for assessing the threat posed by cross-lingual inversion attacks, and the need for continued research in this space.

\section{Experimental Setup}~\label{sec:exp}

\paragraph{English Embeddings} We reproduce the results from~\citet{morris-etal-2023-text} by training inversion models on GTR-base~\citep{ni-etal-2022-large}~\footnote{Huggingface: sentence-transformers/gtr-t5-base} on English dataset. Full results can be found in Appendix~\ref{sec:replication}.

\paragraph{Multilingual Embeddings} 
We use \textbf{T5-base} \citep{raffel2023exploring} as our generation model.
For the multilingual inversion models $\psi$, we train on a state-of-the-art multilingual encoder $\phi$: multilingual-e5-base (\textbf{ME5-base})\footnote{Huggingface: intfloat/multilingual-e5-base}~\citep{wang2022text}, which is a pre-trained transformer based on XLM-R~\citep{conneau2020unsupervised}, and noted to be one of the best performing multilingual models according to MTEB~\citep{muennighoff2023mteb}.

\paragraph{Datasets} 
Previous research~\citep{morris-etal-2023-text} trains inversion models on natural questions and question-answer pairs, such as MSMarco~\citep{bajaj2018ms} and Natural Questions (NQ)~\citep{kwiatkowski-etal-2019-natural}. 
While these datasets are advantageously large, they are limited to English. 
Thus for our experiments, we train and evaluate the multilingual inversion models on \textbf{MTG}, a benchmark suite tailored for multilingual text generation training and evaluation~\citep{chen-etal-2022-mtg}, with parallel samples across languages.
MTG is curated from different domains, including news, daily life, and Wikipedia.
In order to ensure the validity of our experiments, and test generalizability, we exclude the data curated from Wikipedia, as this domain data was already used to train both T5-base and ME5-base models.
For each language, this results in 123k passages (i.e., paragraphs or sections of a document) available for training data. 
We obtain 3-5M sentences for training and 2k each for validation and test in each language using NLTK~\citep{bird-loper-2004-nltk} sentence tokenization.
This is considerably fewer training samples as compared to~\citet{morris-etal-2023-text}, where their GTR-base model was trained on 5M passages from NQ\footnote{The models truncate texts into 32 tokens and 64 tokens, to evaluate how sequence length affects the performance of embeddings inversion. Each passage in NQ is significantly longer than 32 and 64 tokens. To obtain more training data samples from MTG, we implement NLTK sentence tokenization on MTG dataset, resulting in sentences with uneven distribution of tokens length (cf.~Appendix~\ref{appendix:data_distribution}).}.
Meanwhile, we train and evaluate on data in English, French, German and Spanish, noted as MTG-EN, MTG-FR, MTG-DE, and MTG-ES, respectively.
We also compose a 5M-sentence multilingual dataset for training including 1.25M sentences from each language, noted as MTG-MULTI. 
We note that to reproduce the findings presented by \citet{morris-etal-2023-text}, a test set comprising 500 samples was utilized. All reconstruction results are therefore based on 500 samples from the regarding test data.

\paragraph{Metrics} To be comparable with~\citet{morris-etal-2023-text}, we assess model performance using two types of metrics.
First, for text reconstruction, we employ the following word-match metrics: \textit{BLEU}~\citep{post-2018-call}, measuring n-gram similarities between the true and reconstructed text; \textit{ROUGE}~\citep{lin-2004-rouge}, reporting the recall of overlapping words of reconstructed text; \textit{Token F1}, which calculates the multi-class F1 scores between predicted tokens and true tokens, considering each word as a class; and \textit{Exact-match}, representing the percentage of perfectly matching reconstructed texts to the true texts. 
We also compute the \textit{cosine similarity} between the true embedding and the embedding of the reconstructed text in the embedding space of the trained $\phi$.
However, such metrics fall short in terms of evaluating the recovery of the semantic content, especially regarding specific private information. The limitation is particularly evident in cross-lingual settings, for example, where the generated German text conveys similar meaning as the input English text, a nuance that word-match metrics fail to capture (see Figure~\ref{fig:overview}).

\paragraph{Evaluation}
In text generation, exploring the vast space of possible sequences exhaustively is infeasible. Hence, we employ beam search at the sequence level to approximate the sum of immediate text generations. Following~\citet{morris-etal-2023-text}, the inference is conducted greedily at the token level and beam search is employed at the sequence level.
At every stage of correction, a set number $b$ of potential corrections is evaluated. 
For each potential correction, the top $b$ feasible continuations are decoded. 
From the pool of $b \cdot b$ potential continuations, the $b$ unique ones are selected based on their embedding space distance from the reference embedding $e$.

In this study, we analyze inference using varying numbers of correction steps (1, 20, 50, and 100) along with sequence beam widths (sbeam) of 4 and 8.  We explore the impact of evaluation steps in comparison to runtime and observe that evaluation runtime doubles from 50 to 100 steps with sbeam, while the additional performance gains are negligible (see Figure~\ref{fig:runtime_bleu} in Appendix~\ref{appendix:runtime_bleu}). Thus, we report the evaluation results until 50 steps with 8 sbeam.

\begin{table*}[t]
\centering
  \resizebox{\textwidth}{!}{
\begin{tabular}{l|rr|rr|rr|rr|rr|rr|rr}
\hline
  \toprule
  
    &\multicolumn{2}{c}{\textbf{\#Tokens}} & \multicolumn{2}{c}{\textbf{\#Pred Tok.}} & \multicolumn{2}{c}{\textbf{BLEU}} &  \multicolumn{2}{c}{\textbf{ROUGE}} & \multicolumn{2}{c}{\textbf{TF1}} & \multicolumn{2}{c}{\textbf{Exact}} &  \multicolumn{2}{c}{\textbf{COS}}  \\ 
    & \textsc{mono} & \textsc{multi}  & \textsc{mono} & \textsc{multi} & \textsc{mono} & \textsc{multi} & \textsc{mono} & \textsc{multi} & \textsc{mono} & \textsc{multi} & \textsc{mono} & \textsc{multi} & \textsc{mono} & \textsc{multi} \\
  
  \midrule

     \textbf{MTG-EN} &  &  &  &   &  & & &  &  &  &   &  & &  \\ 
        Base (0 Steps) & 32 & 32 & 31.94 & 31.95 & 11.57 & 10.79 & 45.98 & 44.39 & 44.97 & 43.71 & 0 & 0 & 0.9381 & 0.9215 \\ 
        Vec2Text (1 Step) & 32 & 32 & 31.95 & 31.96 & 18.3 & 13.38 & 58.74 & 48.95 & 56.37 & 48.22 & 0.4 & 0.2 & 0.9236 & 0.8637 \\ 
        (20 Steps) & 32 & 32 & 31.99 & 31.98 & 41.48 & 23.72 & 79.05 & 62.53 & 75.15 & 59.74 & 8.8 & 3 & 0.9441 & 0.8433 \\ 
        (50 Steps) & 32 & 32 & 31.99 & 31.97 & 43.05 & 25.27 & 80.2 & 64.14 & 76.29 & 61.39 & 9.4 & 3.2 & \textbf{0.9464} & 0.9296 \\
        (50 Steps + 4 sbeam) & 32 & 32 & 31.99 & 31.98 & 45.87 & 29.89 & 82.7 & 68.17 & 78.24 & 65.27 & 10.8 & 5 & 0.9372 & \textbf{\underline{0.9487}} \\
        (50 Steps + 8 sbeam) & 32 & 32 & 31.98 & 31.98 & \textbf{48.49} & \textbf{32.04} & \textbf{83.51} & \textbf{69.38} & \textbf{79.16} & \textbf{66.67} & \textbf{12} & \textbf{7.4}& 0.9277 & \underline{0.9303} \\ \hline 
        
        \textbf{ MTG-FR}&  &  &  &   &  & & &  &  &  &   &  & &  \\ 
        Base [0 Steps] & 32 & 32 & 32 & 32 & 18.64 & \underline{19.81} & 52.86 & \underline{55.2} & 52.93 & \underline{55.68} & 0 & \underline{0.2} & 0.9408 & \underline{0.9511} \\ 
        Vec2Text (1 Step) & 32 & 32 & 32 & 31.98 & 29.1 & 28.32 & 63.58 & 63.08 & 63.36 & 63.1 & 2.6 & 2 & 0.9655 & 0.9271 \\ 
        (20 Steps) & 32 & 32 & 31.98 & 32 & 62.39 & 58.78 & 84.12 & 81.32 & 83.48 & 81.02 & 36 & 32 & 0.9752 & 0.9492 \\
        (50 Steps) & 32 & 32 & 31.98 & 32 & 64.04 & 60.75 & 85.18 & 83.01 & 84.51 & 82.49 & 36.8 & 33 & 0.9754 & 0.9252 \\
        (50 Steps + 4 sbeam) & 32 & 32 & 32 & 32 & 71.96 & 68.72 & 88.29 & 86.7 & 87.91 & 86.22 & 50.4 & 45.2 & 0.9643 & \textbf{0.942} \\ 
        (50 Steps + 8 sbeam) & 32 & 32 & 32 & 32 & \textbf{74.54} & \textbf{73} & \textbf{89.12} & \textbf{89.38} & \textbf{88.83} & \textbf{88.84} & \textbf{54.4} & \textbf{49.6} & \textbf{0.9757} & \textbf{0.942} \\ \hline

        \textbf{MTG-DE} &  &  &  &   &  & & &  &  &  &   &  & &  \\ 
        Base (0 Steps) & 32 & 32 & 32 & 31.98 & 13.3 & \underline{13.7} & 43.13 & \underline{45.24} & 44.6 & \underline{46.14} & 0 & 0 & 0.9599 & \underline{0.9642} \\ 
        Vec2Text (1 step) & 32 & 32 & 31.93 & 31.98 & 22 & 18.08 & 55.55 & 51.95 & 56 & 52.07 & 1.2 & 0.2 & 0.9699 & 0.9516 \\ 
        (20 Steps) & 32 & 32 & 31.95 & 32 & 56.6 & 41.37 & 80.95 & 70.41 & 79.84 & 69.81 & 30.2 & 16.6 & 0.9573 & 0.9232 \\ 
        (50 Steps) & 32 & 32 & 31.95 & 32 & 57.36 & 43.59 & 82.33 & 72.28 & 81.4 & 71.54 & 30.4 & 17.4 & 0.9687 & 0.9278 \\ 
        (50 Steps + 4 sbeam) & 32 & 32 & 31.98 & 31.98 & 65.79 & 52.48 & 85.84 & 76.7 & 84.56 & 75.75 & 42.4 & 28.2 & \textbf{0.9778} & 0.9321 \\
        (50 Steps + 8 sbeam) & 32 & 32 & 32 & 32 & \textbf{69.5} & \textbf{54.08} & \textbf{87.8} & \textbf{77.57} & \textbf{86.46} & \textbf{76.44} & \textbf{47.4} & \textbf{29.6} & 0.9671 & \textbf{0.9646} \\ \hline

       \textbf{MTG-ES} &  &  &  &   &  & & &  &  &  &   &  & &  \\ 
        Base (0 steps) & 32 & 32 & 31.95 & 32 & 23.21 & \underline{27.09} & 55.15 & \underline{60.54} & 56.75 & \underline{62.07} & 1.6 & \underline{1.8} & 0.938 & \underline{0.9501} \\
        Vec2Text (1 step) & 32 & 32 & 32 & 32 & 35.18 & \underline{36.92} & 66.21 & \underline{68.04} & 67.76 & \underline{68.92} & 8 & \underline{9.6} & 0.9549 & 0.9423 \\ 
        (20 Steps) & 32 & 32 & 32 & 32 & 66.61 & 64.43 & 85.59 & 84.61 & 85.78 & 84.73 & 44.8 & 38.4 & 0.9632 & 0.9563 \\ 
        (50 Steps) & 32 & 32 & 32 & 32 & 67.85 & 65.93 & 86.61 & 85.25 & 86.67 & 85.46 & 45.4 & 38.8 & \textbf{0.9697} & 0.9582 \\ 
        (50 Steps + 4 sbeam) & 32 & 32 & 32 & 32 & 77.29 & 74.52 & 90.41 & 89.45 & 90.47 & 89.23 & 60.8 & 53.6 & \textbf{0.9697} & 0.9515 \\ 
        (50 Steps + 8 sbeam) & 32 & 32 & 32 & 32 & \textbf{80.02} & \textbf{77.72} & \textbf{91.34} & \textbf{90.72} & \textbf{91.54} & \textbf{90.44} & \textbf{65} & \textbf{56.8} & 0.9579 & \textbf{\underline{0.987}} \\ 
        
\bottomrule
\end{tabular}
}
\caption{ 
\textsc{mono} evaluates Text Reconstruction in multiple languages, trained and evaluated on MTG datasets with tokens length 32 in English, French, German, and Spanish, respectively. 
\textsc{multi} evaluates multilingual text reconstruction, trained on MTG-MULTI and evaluated on MTG datasets in the same languages.
The best results across metrics for each language are in \textbf{bold}, with instances where \textsc{multi} outperforms \textsc{mono} \underline{underlined}.
}
\label{tab:mtg-mono-multi}
\end{table*}

\paragraph{Experiments} 
We train an inversion base model and Vec2Text corrector model, as described in Section~\ref{sec:methodology}.
To determine the potential of multilingual embedding inversion attacks, we train base models and Vec2Text models specifically for MTG-MULTI; for cross-lingual attacks, we train these models for each language. In comparison with previous research, we train and evaluate ME5-based inversion models on NQ, i.e., \textsc{me5\_nq}.

We use the Adam optimizer with the learning rate of  $2 \mathrm{e}-5$, epsilon of $1 \mathrm{e}-6$, and 1000 warm-up steps at a constant warm-up schedule. Each base and corrector model is trained for 100 epochs. Due to the prohibitive computational resources needed for training inversion models, we limit each model to a single training run.
For inversion models, we use a batch size of 512, while corrector models, trained on data with 32 tokens, have a batch size of 256. Batch sizes are halved for models trained on data truncated to 64 tokens~\footnote{The more detailed settings for hyper-parameters are illustrated in the GitHub repository.}.
All models are trained on 4 AMD MI250 GPUs with distributed training.
Under these circumstances, training our slowest model takes about 8 days.

\section{Attacking Multilingual Language Models}

To explore the potential of multilingual embedding inversion, we train ME5-base embedder on MTG data in English, German, French, and Spanish, i.e., \textsc{me5\_en}, \textsc{me5\_fr}, \textsc{me5\_de} and \textsc{me5\_es}, respectively, and the composed multilingual dataset of four languages, i.e., \textsc{me5\_multi}, and test on each language for both settings, see results in Table~\ref{tab:mtg-mono-multi}.
To simulate more realistic attacks, we conduct thorough cross-domain evaluation (cf. Appendix~\ref{sec:cross_domain}).

\subsection{Multilingual Text Reconstruction}

\paragraph{Monolingual Text Reconstruction in Multiple Languages}
We observe that the BLEU score for each language peaks by 50 steps correction with 8 sbeam.
Moreover, Spanish models outperform the others in terms of the word-match metrics across correction steps, achieving $80.02$ on BLEU with $65\%$ of exact match. Despite having a larger volume of data compared to other languages, the English model unexpectedly performs the worst across various metrics, as illustrated by the training data distribution in Appendix~\ref{appendix:data_distribution} Figure~\ref{fig:sentence_length}. However, we show in Appendix~\ref{sec:translationese}, the evaluation of round-trip translated English test data indicates no evidence of translationese effect. Additionally, experiments and results for embedding inversion over Finnish and Hungarian can be found in Appendix~\ref{multi_sbert}, providing additional insights to the problem of multilingual Vec2Tex, beyond high-resource Romance and Germanic languages. 
There, we observe sub-par performance for text reconstruction (see: Table~\ref{tab:fin_hun_sbert} of Appendix~\ref{multi_sbert}), highlighting the need to study a wider variety of languages in the future.

\begin{table*}[t]
\centering
  \resizebox{0.9\textwidth}{!}{
    \begin{tabular}{llrr}
    \hline 
\textbf{Step}  & \textbf{Text} & \textbf{BLEU} & \textbf{COS} \\\hline 
Input &	\textbf{ford urged to recall 1.3 million suvs over exhaust fumes} &   &  \\
Step 1	&ford urged to recall \colorbox{yellow}{fumes} \colorbox{pink}{from} 1.3 million suvs &	39.94	& \underline{0.8056}\\
Step 2&	ford urged to recall 1.3 million suvs \colorbox{pink}{from oversowing} fumes &66.06 &	0.9514 \\
Step 3	&ford urged to recall 1.3 million suvs \colorbox{pink}{omitted} fumes	&67.17	&0.8764 \\
Step 4	&ford urged to recall 1.3 million suvs \colorbox{pink}{overfuming} fumes &67.17	& 0.8484\\
Step 5&	ford urged to recall 1.3 million suvs \colorbox{pink}{of} exhaust fumes	&70.71&	\textbf{0.9656}\\
Step 6	& \colorbox{green}{ford urged to recall 1.3 million suvs over exhaust fumes}	& \textbf{100}	& 0.9653\\\hline

     Input & \textbf{ford wird aufgefordert 1,3 millionen suvs wegen abgasen zurückzurufen} &  &  \\ 
     
   Step 1  &  ford \colorbox{pink}{ist auf} 1,3 millionen suvs \colorbox{pink}{zurückgefordertgas abgerufen}& 19.49 & \underline{0.8704} \\
   Step 2 & ford \colorbox{pink}{ist auf} 1,3 millionen suvs \colorbox{pink}{in abgas zurückgefordert} & 19.07 & 	0.8911\\
Step 3 &  ford \colorbox{pink}{ist von} 1,3 millionen suvs wegen \colorbox{pink}{abgas zurückgerufen} & 	31.56& 	0.9592\\
 Step 4 & ford \colorbox{pink}{ist angerufen, dass} 1,3 millionen suvs wegen \colorbox{pink}{abgas zurückgerufen werden}	& 22.42& 	0.9376\\
Step 5 & ford wird aufgefordert\colorbox{pink}{,} 1,3 millionen suvs \colorbox{pink}{aufgrund von abgas} zurückzurufen	& 24.38& 	0.9598\\
Step 6 & ford wird aufgefordert 1,3 millionen suvs wegen \colorbox{pink}{abgas} zurückzurufen	& 75.06	& 0.8906\\
Step 7 & \colorbox{green}{ford wird aufgefordert 1,3 millionen suvs wegen abgasen zurückzurufen} & 	\textbf{100} & 	
 \textbf{0.9872}\\\hline 
    \end{tabular}
    
    }
    
    \caption{Qualitative Analysis of Reconstructing Multilingual Parallel Texts in English and German using \textsc{me5\_multi}. \textbf{Step} are the correction steps from Step 1 (initial hypothesis) to Step 6/7 for the correct inversions. The colored boxes indicate \colorbox{yellow}{misplaced tokens}, \colorbox{pink}{wrong tokens}, and \colorbox{green}{exact matches}. The best results for metrics are in \textbf{bold}. Initial cosine similarity is \textit{underlined}.}
     \label{multi_en_de_qualitative}
\end{table*}

\paragraph{Multilingual Text Reconstruction Without Prior Knowledge of Language}
To evaluate the potential of multilingual text inversion without prior knowledge of the target language, we train inversion models on MTG-MULTI.
As shown in Table~\ref{tab:mtg-mono-multi}, \textsc{me5\_multi} base model outperforms (underlined) or matches the performance of monolingual base models across languages. 
Despite each language in MTG-MULTI having a quarter of the data volume compared to its monolingual counterpart, overall performance remains comparable, particularly evident for French and Spanish. 
For Spanish, \textsc{me5\_multi} slightly outperforms in word-match metrics than \textsc{me5\_es} also for Vec2Text model by 1 step correction. 
Across languages, the initial (base model) cosine similarities of the \textsc{me5\_multi} exceed those of its monolingual counterparts, except for English.

Moreover, we conduct qualitative analysis on text reconstruction using \textsc{me5\_multi} on parallel samples, in Table~\ref{multi_en_de_qualitative} and~\ref{tab:multi_es_fr} (cf. Appendix~\ref{sec:qualitative_analysis}). 
Overall, the lower the cosine similarity of Step 1, the fewer steps the model needs to generate the exact match.
These phenomena suggest that (i) high monolingual data volume is not the sole determinant of high-performing base and 1-step Vec2Text models in both monolingual and multilingual settings, 
(ii) multilingual training yields closer embeddings of reconstructed and target texts in the embedding space, and 
(iii) the optimization approach utilizing cosine similarity is not as effective for multilingual training compared to monolingual.

\subsection{Cross-lingual Text Reconstruction}\label{sec:cross-lingual}

Cross-lingual text reconstruction assumes no prior knowledge of the target language, and thus the embedder $\phi$ is trained on a different source language than the target text for evaluation. 
To investigate the potential of this scenario, we conduct cross-lingual evaluation on all the monolingual models, 
the results on in-domain MTG are reported in Table~\ref{tab:crosslingual_mtg}.

We observe that ME5-base models trained on both NQ and MTG datasets have a tendency to decode texts, for example $\hat{x}$, in the language of training data, e.g., $l_s$, given the target text $x$ which is in a different language, e.g., $l_t$. 
However, $\hat{x}$ could convey the same information in another language, but current word-match metrics are not able to capture this. Thus the privacy leakage still exists. 

For example, the \textsc{me5\_de} model inverts the following German sentence into English:
\begin{itemize}[noitemsep,topsep=1pt]
    \item \textbf{Generated German} report: trump einmal fragte damals fbi director andrew mccabe während seiner 2016-vote
    \item \textbf{AdTrans English} report: trump once asked fbi director andrew mccabe during his 2016-vote
    \item \textbf{Target English} report: trump once asked then-acting fbi director andrew mccabe about his 2016 vote
\end{itemize}
In this case, the model incorrectly generates ``während'' (during) rather than ``about''; otherwise, the generated text is close in meaning with the target English text. 
The information leakage would not be properly captured with the current metrics evaluated on the German text.
Appendix~\ref{sec:qualitative_analysis} Table~\ref{tab:crosslingual_qualitative} shows further qualitative examples for adding AdTrans to aid evaluation in cross-lingual settings.

Finally, for in-domain evaluation, performance improves across cross-lingual settings, as demonstrated in Table~\ref{tab:crosslingual_mtg}. 
Moreover, as shown in Appendix~\ref{sec:cross_domain} Table~\ref{tab:crosslingual_eval_all}, performance is enhanced across models across domains for each language, except for the GTR-base model. 
Notably, the AdTrans strategy proves particularly effective for multilingual based LMs.

\begin{table}[ht!]
 \resizebox{\columnwidth}{!}{
  
    \begin{tabular}{l|cccc}
    \hline 
             & MTG-EN & MTG-FR & MTG-DE  & MTG-ES \\\hline
     \textbf{ \textsc{me5\_en}}  & & & &\\
        Base &  -  & 3.2 (0.9132) & 3.71 (0.8945) & 3.1 (0.9068)  \\
        Vec2Text & - & 4.62 (0.9421) & 5.61 (0.9474)  & 4.33 (0.911)  \\
        AdTrans & -  & \textbf{12.4 ($\uparrow$168.08\%)} & \textbf{6.72 ($\uparrow$19.75\%)} & \textbf{12.38 ($\uparrow$185.79\%)}\\

        \hline 

        \textbf{ \textsc{me5\_fr}} & & & &\\
        Base  &  3.3 (0.9176)  & -& 2.97 (0.9038)  & 4.52 (0.9206) \\ 
        Vec2Text  &  5.36 (0.9235)  & - & 4.26 (0.9431)  &  5.94 (0.9241)\\ 
        AdTrans & \textbf{7.25 ($\uparrow$37.71\%)} & -  &   \textbf{6.35  ($\uparrow$49.47\%)}  &  \textbf{13.7 ($\uparrow$126.79\%)}\\
        
        \hline

        \textbf{\textsc{me5\_de}} & & & &\\
        Base &   3.99 (0.8902)  & 2.96 (0.9082)) & -  & 2.73 (0.9224) \\
        Vec2Text &  8.13 (0.9223) & 4.54 (0.9223)  &  -  &  4.61 (0.9163)\\
        AdTrans & \textbf{9.61 ($\uparrow$18.19\%)} & \textbf{10.37 ($\uparrow$128.62\%)} & -   &  \textbf{11.01 ($\uparrow$138.91\%)}  \\

     \hline 

        \textbf{\textsc{me5\_es}} & & & &\\
        Base &  3.31 (0.9186) & 3.96 (0.9035) &2.67 (0.8958)  & - \\
        Vec2Text &  4.71 (0.9223) &5.13 (0.8699) &3.97 (0.9460)  & - \\ 
        AdTrans & \textbf{5.91 ($\uparrow$25.51\%)}   & \textbf{9.57 ($\uparrow$86.56\%)}  & \textbf{5.56 ($\uparrow$39.89\%)} &  - \\
        \hline 
        
\end{tabular}}
    \caption{
    Cross-lingual evaluation with BLEU score and cosine similarity (in brackets) for Base and Vec2Text models with 50 correction steps and 8 sbeam. BLEU scores and their growth (in brackets) compared with Vec2Text models are reported with AdTrans. 
    $\uparrow$ and $\downarrow$ denote performance gains and losses respectively. The best BLEU results are in \textbf{bold}.
    }
    \label{tab:crosslingual_mtg}
\end{table}

\section{Defending against Inversion Attacks}\label{sec:defense}

To explore defenses against inversion attacks for LMs and compare strategies between monolingual and multilingual models, we investigate the trade-off between retrieval and reconstruction performance. 
Specifically, we apply noise insertion and masking defense to GTR-base and ME5-base using the correction model with 10 steps.
Evaluation is conducted on both BEIR ~\citep{thakur2021beir} (English) and CLIRMatrix~\citep{sun-duh-2020-clirmatrix} (cross-lingual), observing the mean NDCG@10 measures retrieval across 12 tasks (full results in Appendix~\ref{app:full_defense}).

\begin{figure}[tb]
    \centering
        \includegraphics[width=\linewidth]{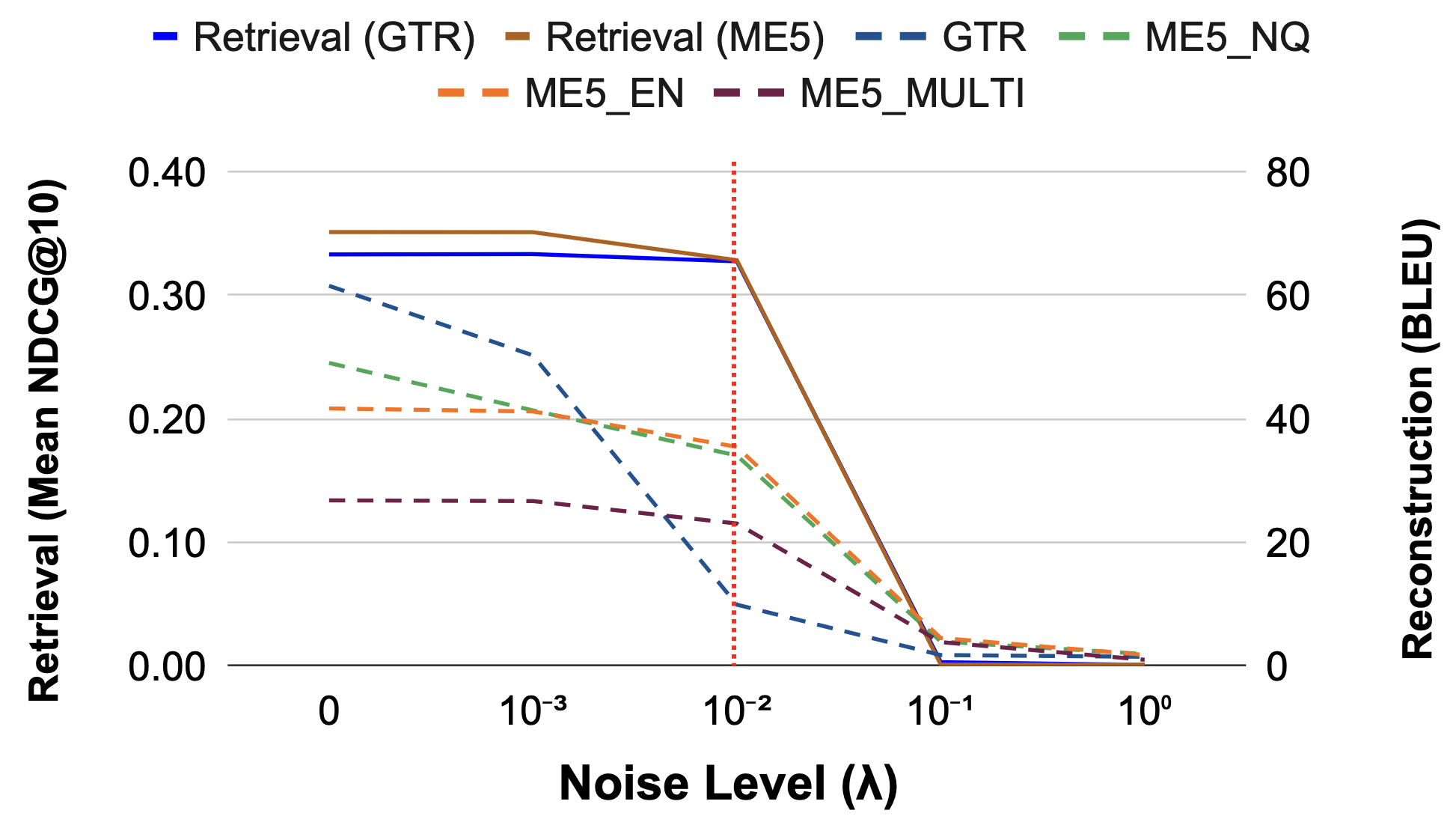}
        \vspace{0.2cm}
        \includegraphics[width=\linewidth]{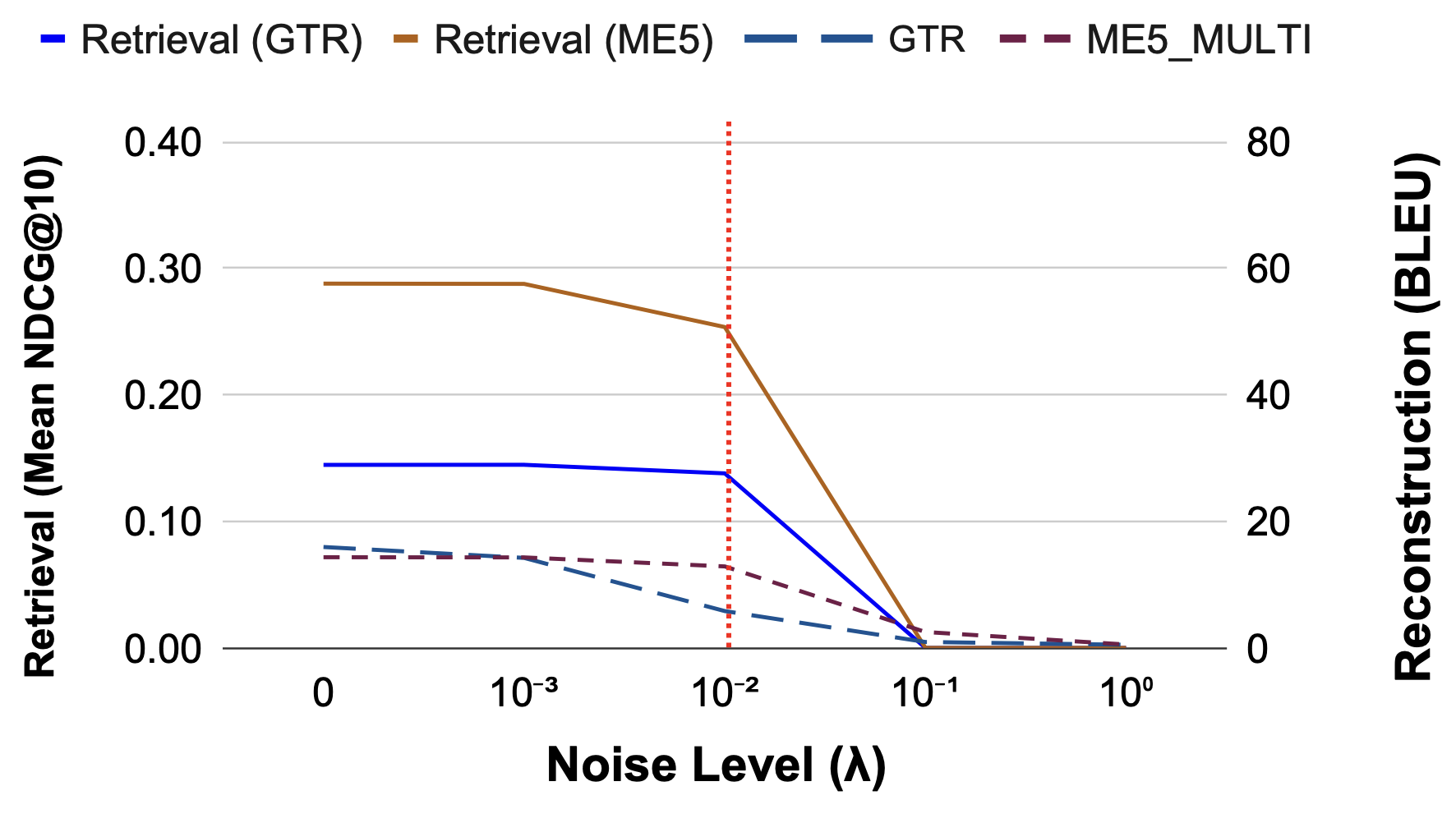}

    \caption{Retrieval and Reconstruction performance \textbf{across varying levels of noise injection} with monolingual (GTR-Based) and multilingual (ME5-Based) language models on BEIR (top) and CLIRMatrix (bottom) datasets. The \textcolor{red}{red dotted lines} indicate the noise level at which the disparity of efficacy of defense between monolingual and monolingual embeddings emerges.  }
    \label{fig:gaussian_defense}
\end{figure}

\begin{figure}[tb]
    \centering
    \includegraphics[width=\linewidth]{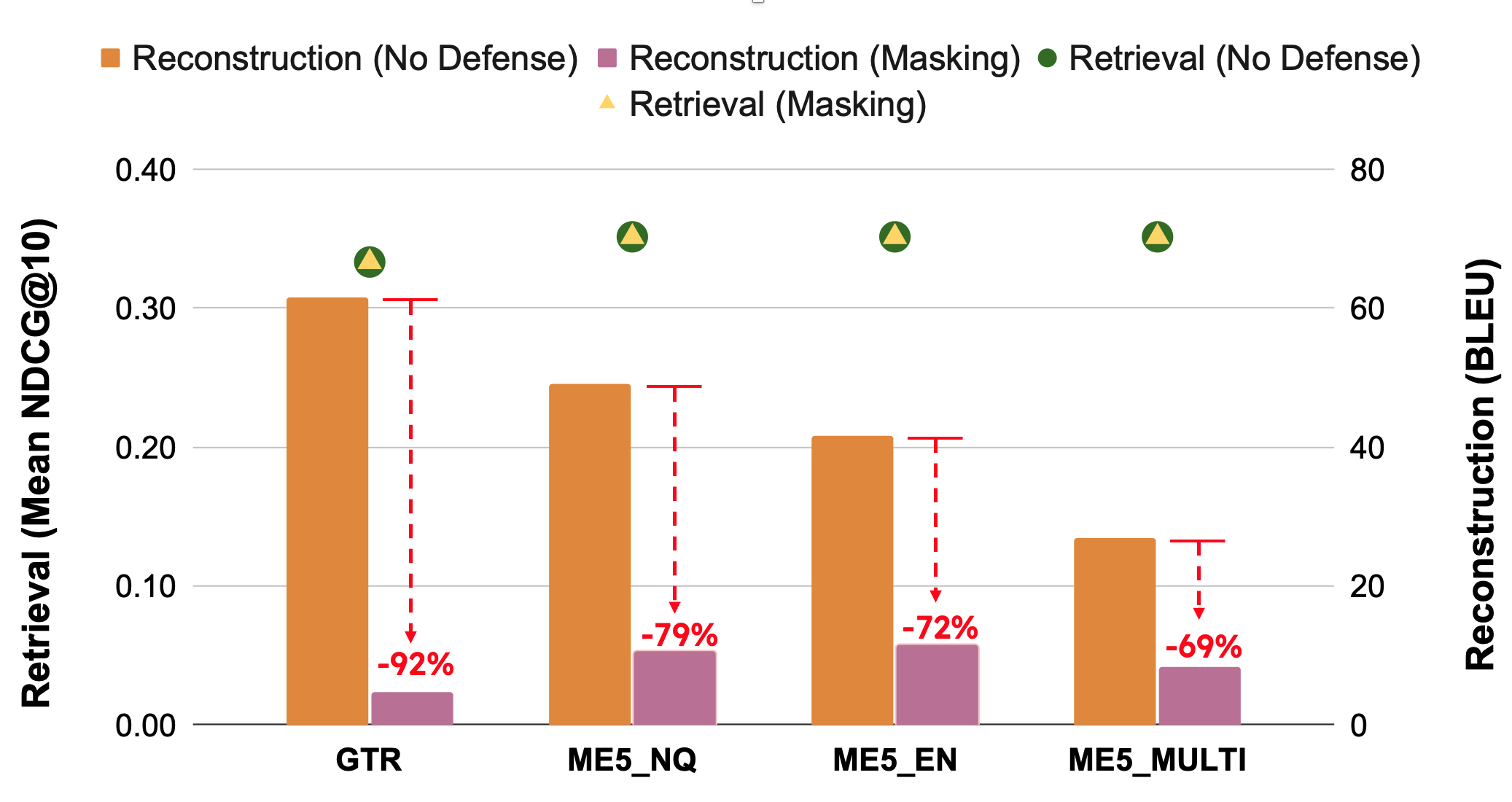}
    \vspace{0.2cm}
    \includegraphics[width=0.7\linewidth]{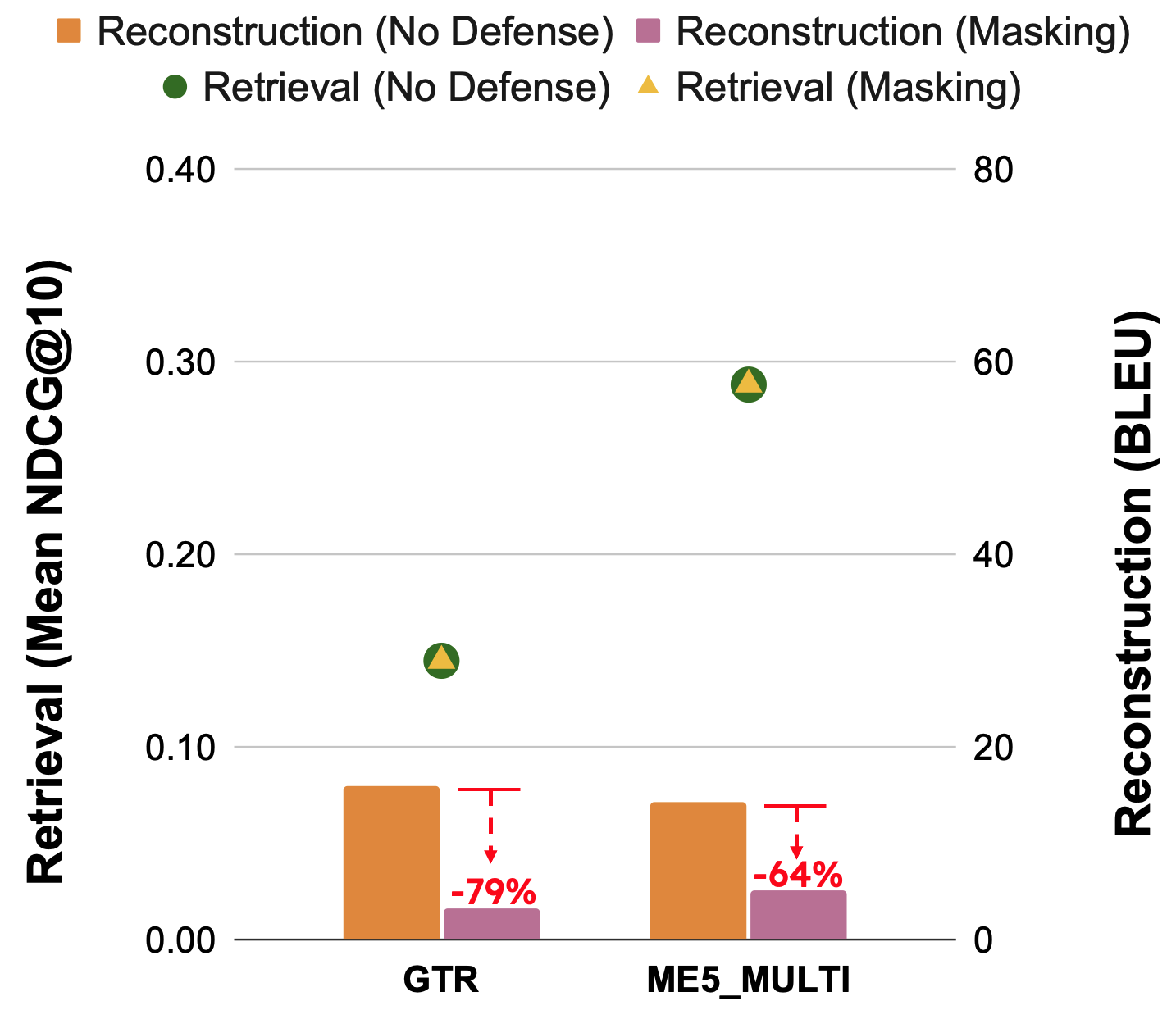}
    \caption{Retrieval and Reconstruction performance with \textbf{masked} monolingual (GTR-Based) and multilingual (ME5-Based) language models on BEIR (top) and CLIRMatrix (bottom) datasets. The \textcolor{red}{red dashed lines} indicate the performance drop in percentage. }
    \label{fig:masking_defense}
\end{figure}

\paragraph{Inserting Noise}
Simple noise insertion (detailed in Appendix~\ref{app:gaussian_noise}) effectively guards monolingual LMs against inversion attacks~\citep{morris-etal-2023-text},
which is confirmed by our experiments, demonstrating that adding noise can defend against such attacks while preserving embedding utility, as depicted in Figure~\ref{fig:gaussian_defense}. 

With a noise level of $\lambda=10^{-3}$, retrieval performance is preserved for both \textsc{gtr} and \textsc{me5} across BEIR and CLIRMatrix. While there is a drop on reconstruction with \textsc{gtr} and  \textsc{me5\_nq} on BEIR by 20\%, there is no change with \textsc{me5\_en} on BEIR and \textsc{me5\_multi} on both BEIR and CLIRMatrix. 

At the noise level $10^{-2}$, reconstruction performance with \textsc{gtr} drastically drops to 16\% of the original BLEU on BEIR and 36\% on CLIRMatrix. In contrast, reconstruction with multilingual LMs consistently maintains over 70\% of the original BLEU, particularly with \textsc{me5} trained on MTG over 85\%.
Additional noise ($\lambda\geq10^{-1}$) damages significantly both retrieval and reconstruction performances. 
This notable disparity between retrieval and reconstruction performance on \textsc{gtr} ($\lambda=10^{-2}$) implies the efficacy of the noise insertion defense primarily on monolingual LMs rather than multilingual ones.

\paragraph{A Frustratingly Simple Masking Defense}
To enhance the security of LMs, we propose a simple defense method, achieved by masking the first dimension of the embeddings with the encoding of the target language $l_t$. 
We use an iterator to encode each language as an identifier, denoted as $id_{t} \in \mathbb{R}$. 
The masked embedding model is defined as following:
\begin{equation}\label{eq:masking}
    \phi_{masking}(x) = vec([id_{t}, vec(\phi_{i}(x))_{1\leq i \leq n}])
\end{equation}
given $\phi(x)=vec(\phi_{i}(x))_{ 0\leq i \leq n}$ where $x$ is the input text, $n$ is the dimension of the embedding $\phi(x)$ and $n\in \mathbb{N}$.

We implement this simple masking defense on both GTR-base and ME5-base models. As depicted in Figure~\ref{fig:masking_defense},
while retrieval performance remains unaffected\footnote{The performance of text reconstruction on CLIRMatrix dataset with~\textsc{gtr} is largely conflated by its superiority in reconstructing English documents (details in Appendix~\ref{app:full_defense}).} across all models, reconstruction markedly declines for both monolingual and multilingual LMs across the retrieval benchmarks, with a notable drop by 92\% with \textsc{gtr} on BEIR and 79\% on CLIRMatrix, and over 64\% drop for all multilingual models. 
The is a simple yet effective defense against inversion attacks for both monolingual and multilingual LMs, while fully preserving utility in retrieval tasks.

\section{Conclusion}

While previous works on embedding inversion attacks focus exclusively on English, we present the first work on multilingual and cross-lingual embedding inversion.
Notably, we uncover that multilingual models can be \textit{more} vulnerable than monolingual models, under certain conditions.
Importantly, traditional defense tailored for monolingual models prove ineffective in guarding multilingual models. 
Thus we propose a more robust defense applicable to both monolingual and multilingual ones. 
Additionally, our preliminary experiments over moderately-resourced Uralic languages further stresses the importance of expanding the scope of future works in embedding inversion studies, to include a more diverse set of languages.
In summary, our work advocates for a multilingual approach to LLM and NLP security as an entirety.

\section*{Limitations}
\paragraph{Computing Resources}
A core limitation of this work is the computationally intense experiments, requiring in the area of 25,000 GPU computing hours.
While expanding this research direction to more languages will further increase this expense, we advocate for ensuring that languages other than English are not left behind in terms of NLP security.

\paragraph{Data Contamination} 
Pre-trained LMs are often trained on massive web-based datasets, resulting in a high likelihood that a given model has already seen commonly used benchmark datasets \cite{dodge-etal-2021-documenting}. Indeed, most wide-used LMs are trained on massive datasets like the C4 Common Crawl~\footnote{\url{https://commoncrawl.org}} web scrape, including OpenAI's GPT models~\citep{radford2019language,brown2020language}, Meta AI's RoBERTa~\citep{liu2019roberta} and LLaMAs~\citep{touvron2023llama}, Google AI's BERT~\citep{devlin2018bert}, and EleutherAI's GPT-Neo~\citep{black2022gpt} and GPT-J~\citep{mesh-transformer-jax}.
In this work, we utilize models including T5-base, ME5-base and GTR-base, which are all trained on massive public domain datasets, resulting in a likely overlap of training data. For example, initialized from T5, GTR-base is trained on NQ dataset, which is again used as training data for text reconstruction by~\citet{morris-etal-2023-text}; ME5-base and T5-base overlaps in C4 and Wikipedia. In an attempt to mitigate data contamination, we exclude Wikipedia from the MTG dataset. However, staving off data contamination entirely is nearly infeasible when utilizing open-sourced pre-trained large LMs. This limitation is the focus of several previous works~\citep{brown2020language,dodge2021documenting,magar-schwartz-2022-data, jacovi-etal-2023-stop}.

\paragraph{Number and Diversity of Languages} In this study, we extensively experiment on multilingual and cross-lingual inversion security focused on four Romance and Germanic languages, which are also high-resource languages in NLP. 
Still, this means that this work lacks the extensive linguistic diversity needed to understand how embedding inversion attacks affect massively multilingual models, or lower-resourced languages.
To this end, we include some preliminary experiments for inverting multilingual sentence BERT in two Uralic languages, i.e., Finnish and Hungarian. Ultimately, we advocate for more extensive research with a wider sample of languages in various language families.

\section*{Ethics Statement}
This work explores attacks on multilingual embedding models.
Our intent with this research is to shed light on the vulnerabilities of languages other than English, aiming to encourage the community to include more languages in NLP security work.
While there is potential for misuse by malicious actors, as with many works in NLP security, we mitigate harm by including an effective countermeasure to the attack presented in the paper.
Still, it is important to stress that embedding inversion presently represents a substantial threat.
To this end, the LMs examined in this paper are open-source models, and such that this work does not constitute an imminent threat to EaaS providers, who are likely using private models. 
Finally, we do not knowingly experiment with any truly sensitive data, ensuring that no real-world harm is caused by the work carried out in this paper.

\section*{Acknowledgements}
All authors of this paper are funded by the Carlsberg Foundation, under the Semper Ardens: Accelerate programme (project nr.~CF21-0454). 
We are furthermore grateful to the support of the AAU AI Cloud, and to DeiC for allocating us computing resources on the LUMI cluster (project nr.~DeiC-AAU-S5-412301).
We thank Sighvatur Sveinn Davidsson for setting us up with this access, and for his diligence in assisting with the problems in the experimental infrastructure, in addition to the LUMI user support for their very prompt answers and competence, especially Jing Gong. 
We  further thank Esther Ploeger for her assistance in testing translationese effect for the under-performance of multilingual inversion model in English and Marcell Richard~Fekete for his insightful input in proof-reading the paper.

\bibliography{anthology,custom}
\bibliographystyle{acl_natbib}

\newpage
\appendix

\begin{table*}[ht!]
    \centering
     \resizebox{ \textwidth}{!}{
    \begin{tabular}{l|rr|rr|rr|rr|rr|rr|rr}
    \toprule
    &\multicolumn{2}{c}{\textbf{\#Tokens}} & \multicolumn{2}{c}{\textbf{\#Pred Tok.}} & \multicolumn{2}{c}{\textbf{BLEU}} &  \multicolumn{2}{c}{\textbf{ROUGE}} & \multicolumn{2}{c}{\textbf{TF1}} & \multicolumn{2}{c}{\textbf{Exact}} &  \multicolumn{2}{c}{\textbf{COS}}  \\ 
    
   &  \textbf{GTR} & \textbf{ME5} & \textbf{GTR} & \textbf{ME5} & \textbf{GTR} & \textbf{ME5} & \textbf{GTR} & \textbf{ME5} & \textbf{GTR} & \textbf{ME5} & \textbf{GTR} & \textbf{ME5} & \textbf{GTR} & \textbf{ME5}    \\ 
   
  \midrule
        Base (0 Steps) & 32 & 32 & 32 & 32 & 27.18 & \underline{28.77} & 62.86 & \underline{63.68} & 63.74 & \underline{65.9} & 0.4 & 0.4 & 0.8793 & \underline{0.9738} \\ 
        Vec2Text (1 Step) & 32 & 31 & 32 & 32 & 48.62 & 47.92 & 78.39 & 77.03 & 78.44 & 78.35 & 8 & 4.8 & 0.921 & \underline{0.9588} \\
        (20 Steps) & 32 & 32 & 32 & 32 & 83.30 & 74.47 & 95.12 & 89.57 & 95.11 & 90.3 & 58 & 21.8 & 0.9862 & 0.992 \\ 
        (50 Steps) & 32 & 32 & 32 & 32 & 84.31 & 75.03 & 95.49 & 89.76 & 95.6 & 90.56 & 58.4 & 21.8 & 0.9862 & 0.992 \\
        (50 Steps + 4 sbeam) & 32 & 32 & 32 & 32 & 90.18 & 78.87 & 97.26 & 91.11 & 97.15 & 91.55 & 74.4 & 32.6 & 0.9853 & 0.9902 \\ 
        (50 Steps + 8 sbeam) & 32 & 32 & 32 & 32 & 92.44 & \textbf{80.86} & \textbf{97.76} & \textbf{91.89} & 97.78 & \textbf{92.42} & \textbf{82} & \textbf{35} & \textbf{0.9921} &  \textbf{\underline{0.9926}} \\ 
        (100 Steps) & 32 & 32 & 32 & 32 & \textbf{92.45} & 80.82 & 97.75 & 91.83 & \textbf{97.79} & 92.37 & 82 & 35 & 0.9921 & \underline{0.9926} \\ 
        (100 Steps + 4 sbeam) & 32 & 32 & 32 & 32 & 90.17 & 78.82 & 97.25 & 91.11 & 97.15 & 91.53 & 74.4 & 32.8 & 0.9824 & \underline{0.9902} \\
        (100 Steps + 8 sbeam) & 32 & 32 & 32 & 32 & 92.45 & 80.82 & 97.75 & 91.83 & 97.79 & 92.37 & 82 & 35 & 0.9921 & \underline{0.9926} \\

 \bottomrule
    \end{tabular}}
    \caption{Evaluation of English Text Reconstruction. The best performances for each model reached in the earliest stages are in \textbf{bold}. The \underline{underlined} results are where ME5-base model outperforms GTR-base model.
    }
    \label{tab:replication}
\end{table*}

\section{Training Data Distribution }~\label{appendix:data_distribution}

\begin{figure}[htb]
    \centering
    \includegraphics[width=\linewidth]{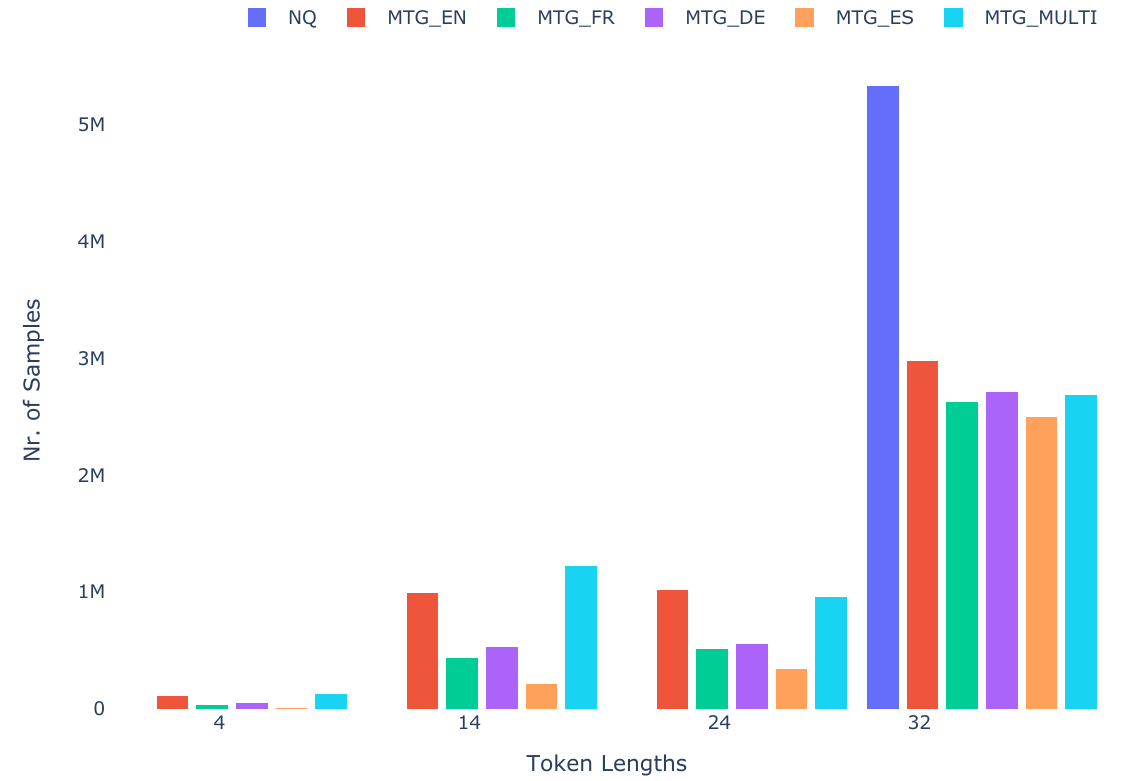}
    \caption{The Distribution of the training data for models with the maximal token length of 32.  }
    \label{fig:sentence_length}
\end{figure}

In the MTG datasets, English texts are sourced from various origins, while German, Spanish, and French texts are translated from English using machine translation and manually validated~\citep{chen-etal-2022-mtg}.
These languages exhibit diverse morphologies, leading to variations in sentence lengths and the number of sentences post-tokenization across languages. 
Additionally, the NQ dataset is included to reproduce findings from prior research \citep{morris-etal-2023-text} and to assess the cross-domain and cross-lingual performance of the text reconstruction task. 
The NQ dataset predominantly comprises English data, with Wikipedia passages included without tokenization, resulting in all training data from NQ having 32 tokens.

\section{Monolingual English Text Reconstruction}\label{sec:replication}


To have a proof of concept, we successfully reproduce and replicate the experiment from~\citet{morris-etal-2023-text}, by training inversion models using GTR-base and ME5-base as embedders on the NQ dataset, noted as \textsc{gtr} and \textsc{me5\_nq}.

The results for reconstructing English texts are shown in Table~\ref{tab:replication}, evaluated with correction steps (1, 20, 50, 100) combined with beam search (4 and 8 sbeam).
The base and 1-Step Vec2Text model trained on ME5-base have a performance on par with GTR-base. 
Moreover, the text embeddings trained on ME5-base are closer in embedding space than embeddings trained on GTR-base, i.e., with higher cosine similarities. 

While, with more steps of correction and sbeam, the performance is boosted to 92.45 on BLEU with $82\%$ exact match for \textsc{gtr}, while the best performance for \textsc{me5\_nq} is 80.86 on BLEU with $35\%$ exact match.
The performance difference could be due to the fact that the underlying GTR-base is t5-based model, the same structure as the generation model $\psi$. 

However, utilizing ME5-base sets up a more realistic attack scenario of black-box embedding inversion, as the structure of the embedder $\phi$ is unknown.
Both models are furthermore evaluated with cross-domain English text reconstruction. 
Similarly, \textsc{gtr} outperforms \textsc{me5} after 50 correction steps with sbeam 8, see Table~\ref{tab:crossdomain-en} in Appendix~\ref{sec:cross_domain}.

\section{Runtime vs. BLEU scores }\label{appendix:runtime_bleu}
\begin{figure}[h!]
    \centering
    \includegraphics[width=\linewidth]{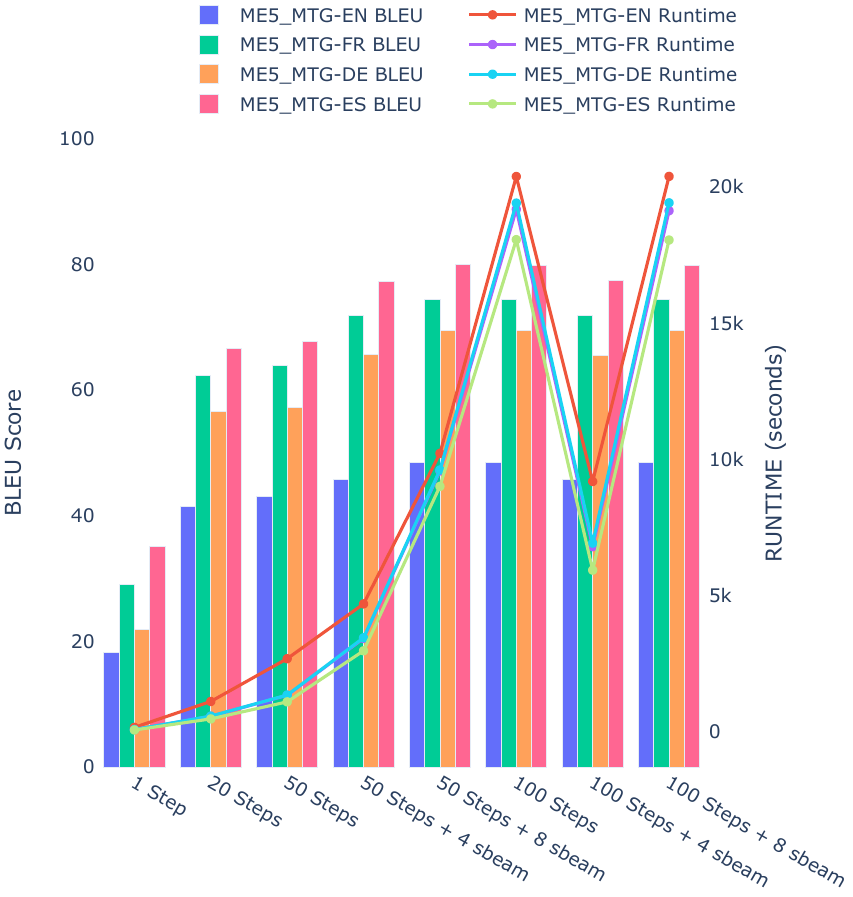}
    \caption{BLEU scores vs. Runtime by Evaluation for Inversion Models in English, French, German and Spanish.}
    \label{fig:runtime_bleu}
\end{figure}

The evaluation of Vec2Text models is expensive in terms of time and computation.
In order to search for the optimal runtime and performance trade-off, Figure~\ref{fig:runtime_bleu} shows BLEU scores at each step and the lines represent the trend for runtime for the monolingual models. 
The best trade-off points are at the correction step of 50 with 8 sbeam for all the models, while 100 steps takes more than double the time achieving similar performance. The full results are in Table~\ref{tab:mtg-mono-multi} and \ref{tab:mtg-mono}.
Until correction step 50 with 8 sbeam, performance increases steadily, and the trend is generally aligned with cosine similarity. 
As a result, we evaluate the subsequent models until correction step 50 with 8 sbeam.

\begin{table}[ht!]
\centering
  \resizebox{\linewidth}{!}{
\begin{tabular}{l|rrrrrrr}
\hline
  \toprule

    & \textbf{\#Tokens} & \textbf{\#Pred Tok.} & \textbf{BLEU} & \textbf{ROUGE} & \textbf{TF1} & \textbf{Exact} & \textbf{COS}\\
  
  \midrule
  \textbf{MTG-EN} & & & & & & & \\ 
        (100 Steps) & 32 & 31.98 & 48.53 & 83.51 & 79.12 & 12 & 0.9277 \\ 
        (100 Steps + 4 sbeam) & 32 & 31.99 & 45.9 & 82.71 & 78.24 & 10.8 & 0.9372 \\
        (100 Steps + 8 sbeam) & 32 & 31.98 & 48.53 & 83.51 & 79.12 & 12 & 0.9277 \\\hline
        \textbf{ MTG-FR}& & & & & & & \\ 
        (100 Steps) & 32 & 32 & 74.44 & 89.1 & 88.77 & 54.4 & 0.9757 \\ 
        (100 Steps + 4 sbeam ) & 32 & 32 & 71.93 & 88.26 & 87.89 & 50.4 & 0.9643 \\ 
        (100 Steps + 8 sbeam) & 32 & 32 & 74.44 & 89.1 & 88.77 & 54.4 & 0.9757 \\ \hline 
        \textbf{MTG-DE} & & & & & & & \\ 
        (100 Steps) & 32 & 32 & 69.55 & 87.8 & 86.47 & 47.4 & 0.9791 \\ 
        (100 Steps + 4 sbeam) & 32 & 31.98 & 65.61 & 85.73 & 84.46 & 42.2 & 0.9778 \\
        (100 Steps + 8 sbeam) & 32 & 32 & 69.55 & 87.8 & 86.47 & 47.4 & 0.9791 \\ \hline
        \textbf{MTG-ES} & & & & & & & \\ 
        (100 Steps) & 32 & 32 & 79.96 & 91.21 & 91.43 & 65 & 0.9579 \\
        (100 Steps + 4 sbeam) & 32 & 32 & 77.48 & 90.52 & 90.56 & 60.8 & 0.9697 \\
        (100 Steps + 8 sbeam) & 32 & 32 & 79.96 & 91.21 & 91.43 & 65 & 0.9579 \\

\bottomrule
\end{tabular}
}
\caption{The evaluation of Text Reconstruction in multiple languages, with the models trained and evaluated on MTG datasets with tokens length 32 in English, French, German and Spanish, respectively. The steps are from 100 steps to 100 steps + 8 sbeam.}
\label{tab:mtg-mono}
\end{table}

\section{Inverting Multilingual Sentence BERT Embeddings}~\label{multi_sbert}

We additionally experiment on inverting multilingual sentence BERT in Finnish and Hungarian. The inversion models are trained using the encoder-decoder multilingual T5~\citep{wang2024multilingual} as generation model, and multilingual sentence BERT~\footnote{huggingface: sentence-transformers/distiluse-base-multilingual-cased-v2} is used as the encoder $\phi$. We train models on randomly extracted 1M data samples from CulturaX~\citep{nguyen-etal-2024-culturax-cleaned}~\footnote{huggingface: uonlp/CulturaX}, validated and evaluated on 500 samples, respectfully.  The detailed evaluation results are reported in Table~\ref{tab:fin_hun_sbert}. Interestingly, the corrector model, which converges embeddings with cosine similarity, did not improve text reconstruction for Finnish texts, while it did provide marginal improvement for Hungarian texts. The notably poorer performance in this experiment highlights the complexity of inverting textual embeddings, where model affinity and datasets play crucial roles. For future work, we plan to investigate more extensively how different model architectures and language families influence embedding inversion performance.

\begin{table}[ht!]
    \centering
     \resizebox{0.5\textwidth}{!}{
    \begin{tabular}{l|ccccccc}
           
    \hline 

      & \textbf{\#Tokens}  & \textbf{\#Pred Tok.}   &  \textbf{BLEU} &  \textbf{ROUGE} &     \textbf{TF1}  & \textbf{EXACT}      & \textbf{COS}   \\\hline 
      \textbf{Finnish} & & & & & & & \\
       Base (0 Steps)  & 32 & 31 & \textbf{7.69} &  \textbf{0.24} & \textbf{0.27} &   \textbf{0.014} &  \textbf{0.7068}  \\
        Vec2Text (1 Step) & 32 &    0.0  & 0.0  &   0.0  & 0.00    & 0.0  & -0.0562  \\
        (20 Steps) & 32 & 0.0 &  0.0  &  0.0 &  0.0 &    0.0  & -0.0562  \\
        (50 Steps) & 32&  0.0  & 0.0  &   0.0 &  0.0 &  0.0  & -0.0562  \\
        (50 Steps + 4 sbeam) & 32 &  31  & 0.01    & 0.0  & 0.13   &  0.0  &-0.0166    \\
        (50 Steps + 8 sbeam) &32  & 8.0 & 0.03    &0.0  & 0.14   & 0.0 & 0.0034     \\
            \hline 

        \textbf{Hungarian} & & & & & & & \\
       Base (0 Steps)  & 32 & 31 &  6.74 & 0.31 & 0.30 &  0.002 &  0.6834  \\
        Vec2Text (1 Step) &  32 &  31 &   7.15 &   0.32 &  30.52  &  \textbf{0.2} & 0.7220   \\
        (20 Steps) & 32  & 31 & 7.35  & 0.32  &  30.99   &  0.2 &  0.7170    \\
        (50 Steps) & 32  & 31 & 7.37  & 0.32  & 31.04   &  0.2   &0.7170   \\
        (50 Steps + 4 sbeam) & 32 & 31 &  7.95 &  0.33 &   \textbf{31.76}  &  0.0 & 0.7564    \\
        (50 Steps + 8 sbeam) & 32 &  31  & \textbf{8.00} & \textbf{0.33} &  31.28   &  0.0  & \textbf{0.8240}   \\
            \hline 
    \end{tabular}}
        \caption{Inverting Multilingual Sentence BERT textual embeddings in Finnish and Hungarian. The best results for each metric are in \textbf{bold}.}
    \label{tab:fin_hun_sbert}
\end{table}

\section{No Evidence for Translationese Effect}\label{sec:translationese}
In machine translation, there is clear evidence that the presence of translationese in test sets may result in inflated human evaluation scores for MT systems ~\citep{zhang-toral-2019-effect}. To investigate whether our multilingual inversion model's sub-par performance in English is due to the characteristics of translationese in other languages, we implement round trip translation on MTG-EN test data using Spanish as the pivot language with EasyNMT, the translation path is thus English $\rightarrow$ Spanish $\rightarrow$ English. Then the evaluation of the multilingual inversion model is done on the round-trip translated English test set, the result is shown as in Table~\ref{tab:translationese}. Compared to evaluation on MTG-EN test set, as shown in Table~\ref{tab:mtg-mono-multi}, the performance of translated English test set is about 30 on BLEU worse at each stage of corrections. 
The hypothesis of the translationese effect on the difference of the performances can therefore be rejected. 

\begin{table}[ht!]
    \centering
    \resizebox{0.5\textwidth}{!}{
    \begin{tabular}{l|ccccccc}
    \hline 
         & \textbf{\#Tokens}  & \textbf{\#Pred Tok.}   &  \textbf{BLEU} &  \textbf{ROUGE} &     \textbf{TF1}  & \textbf{EXACT}      & \textbf{COS}   \\\hline 
Vec2Text (1 Step) & 29.59 & 30.98 & 10.03 & 47.54 & 41.28 & 0 & 0.9046 \\ 
        (20 Steps) & 29.59 & 30.95 & 14.48 & 55.14 & 47.8 & 0.2 & 0.913 \\ 
        (50 Steps) & 29.59 & 30.98 & 15.11 & 56.01 & 48.56 & 0.2 & 0.9261 \\ 
        (50 Steps + 4 sbeam ) & 29.59 & 30.88 & 17.56 & 61.81 & 52.64 & 0.2 & 0.9461 \\ 
        (50 Steps + 8sbeam) & 29.59 & 30.96 & 17.42 & 61.28 & 52.44 & 0.4 & 0.9185 \\ \hline
         
    \end{tabular}}
    \caption{Evaluation of multilingual inversion model on round-trip translated MTG-EN test dataset.}
    \label{tab:translationese}
\end{table}

\section{Text Construction on Tokens Length 64}\label{sec:tokens64_eval}

\begin{table}[ht!]
    \centering
     \resizebox{0.5\textwidth}{!}{
    \begin{tabular}{l|ccccccc}
    \hline 
      &\textbf{\#Tokens} & \textbf{\#Pred Tokens} & \textbf{BLEU} & \textbf{ROUGE} & \textbf{TF1} & \textbf{Exact} & \textbf{COS} \\ \hline
 \textbf{English}  & & & & & & & \\
        Vec2Text (1 Step) & 37.78 & 43.73 & 18.13 & 59.33 & 57.28 & 0.8 & 87.94 \\ 
        (20 Steps) & 37.78 & 41.32 & 38.48 & 78.38 & 74.23 & 10 & 88.75 \\ 
        (50 Steps) & 37.78 & 40.97 & 39.27 & 79.74 & 75.4 & 10.2 & \textbf{92.70} \\ 
        (50 Steps + 4 sbeam) & 37.78 & 40.67 & 45.23 & 81.68 & 77.31 & 14.6 & 89.18 \\ 
        (50 Steps + 8 sbeam) & 37.78 & 40.19 & \textbf{47.29} & \textbf{83.34} & \textbf{78.62} & \textbf{16.6} & 91.09 \\ \hline
        
        \textbf{French} & ~ & ~ & ~ & ~ & ~ & ~ & ~ \\ 
        Vec2Text (1 Step) & 51.61 & 57.23 & 26.45 & 63.58 & 64.03 & 0.8 & 95.07 \\ 
        (20 Steps) & 51.61 & 53.25 & 58.25 & 83.1 & 83.01 & 26.6 & 96.54 \\ 
        (50 Steps) & 51.61 & 52.6 & 59.58 & 83.99 & 83.69 & 26.8 & 96.26 \\ 
        (50 Steps + 4 sbeam) & 51.61 & 52.62 & 64.61 & 86.11 & 86.03 & 37.8 & \textbf{97.26} \\ 
        (50 Steps + 8 sbeam) & 51.61 & 52.54 & \textbf{66.8} & \textbf{86.74} & \textbf{86.44} & \textbf{41.8} & 93.83 \\ \hline

       \textbf{German} & & & & & & & \\
        Vec2Text(1 Step) & 49.75 & 56.09 & 19.65 & 54.58 & 55.19 & 0.2 & \textbf{97.43} \\
        (20 Steps) & 49.75 & 52.62 & 46.11 & 76.1 & 75.3 & 15.6 & 93.98 \\ 
        (50 Steps) & 49.75 & 52.76 & 46.61 & 76.69 & 75.86 & 15.8 & 95.72 \\ 
        (50 Steps + 4 sbeam) & 49.75 & 51.91 & 52.78 & 79.6 & 78.93 & 25.6 & 92.98 \\ 
        (50 Steps + 8 sbeam) & 49.75 & 51.82 & \textbf{55.73} & \textbf{80.87} & \textbf{80.21} & \textbf{30.8} & 94.97 \\ \hline
        
        \textbf{Spanish}  & & & & & & & \\
        Vec2Text(1 Step) & 62.66 & 62 & 26.03 & 64.16 & 65.78 & 0.4 & 97.57 \\ 
        (20 Steps) & 62.66 & 62.23 & 56.07 & 83.53 & 83.7 & 17.4 & \textbf{98.28} \\ 
        (50 Steps) & 62.66 & 62.09 & 56.73 & 84.37 & 84.46 & 17.4 & 97.01 \\ 
        (50 Steps + 4 sbeam) & 62.66 & 61.95 & 64.27 & 86.78 & 87.01 & 29.2 & 95.39 \\ 
        (50 Steps + 8 sbeam) & 62.66 & 61.76 & \textbf{65.57} & \textbf{87.73} & \textbf{87.85} & \textbf{32.8} & 97.36 \\ \hline

    \end{tabular}}
    \caption{The evaluation of Text Reconstruction in multiple languages, with the models trained and evaluated on MTG datasets with maximal token length 64 in English, French, German and Spanish, respectively. The best results across metrics are in \textbf{bold}.}
    \label{tab:eval_64}
\end{table}

We train ME5-base inversion models on MTG datasets with token lengths of 64 in English, French, German, and Spanish, in comparison to 32-token length models. Results in Table~\ref{tab:eval_64} indicate a performance degradation; for instance, the BLEU score for the Spanish inversion model drops by approximately 15 while doubling the number of tokens. This highlights the challenges in this line of research.

\section{Cross-Domain Text Reconstruction}\label{sec:cross_domain}

\paragraph{Cross-Domain English Text Reconstruction} 

\begin{table}[htb]
    \centering
    \resizebox{0.49\textwidth}{!}{
    \begin{tabular}{l|c|c|c}
    \hline 
    & NQ$\rightarrow$MTG-EN   & MTG-EN$\rightarrow$NQ & MTG-MULTI$\rightarrow$NQ \\\hline 
       \textbf{GTR}  &   &  & \\
       Base   &  5.81 (0.7334) & -  & -\\ 
        Vec2Text & \textbf{39.08} (0.9767) &   & \\ \hline 
        \textbf{ME5}  &  & \\
        Base  & \textbf{5.89} (0.9272) &  \textbf{12.35} (0.9154) &   11.63 (0.8894) \\
        Vec2Text  & 26.96   (0.9440) & \textbf{42.90} (0.9789)  &  31.84 (0.9310)  \\\hline
    \end{tabular}
    }

    \caption{Cross-Domain English Text Reconstruction Evaluation, BLEU scores and COS are reported.
    Horizontal comparison on ME5-base models, and vertically on two embedders trained on the same NQ dataset. The Vec2Text models are evaluated by 50 steps of correction with sequence beam search width 8. $\rightarrow$ indicates the cross-domain evaluation direction. For example, NQ $\rightarrow$ MTG-EN indicates that the model is trained on NQ and evaluated on MTG-EN. }
    \label{tab:crossdomain-en}
\end{table}

To evaluate the performance of embedding inversion attacks on out-of-domain dataset in English, the models trained on NQ and MTG-EN are cross-evaluated on both datasets, respectively, as shown in Table~\ref{tab:crossdomain-en}. The results on MTG-EN are similar on BLEU for both base models trained on GTR-Base and ME5-Base, while \textsc{gtr} model outperforms \textsc{me5} by more than 12 on BLEU, and the cosine similarity of reconstructed and true text embeddings are boosted by over $0.24$ . In comparison, the cosine similarity for \textsc{me5} models are not much varied and constantly high ($\geq 0.88$) across stages of evaluations and across domains. 
Additionally, \textsc{me\_en} outperforms \textsc{me\_multi} tested on NQ.

\paragraph{Cross-domain Cross-lingual Text Reconstruction}

Cross-lingual, cross-domain text reconstruction is one of the most challenging scenarios, yet it also represents the most realistic context, with both domain and target language unknown.
As shown in Table~\ref{tab:crosslingual_eval_nq2mtg},
while the AdTrans strategy does not enhance the performance of the GTR-Base inversion model, there is a consistent improvement in performance across datasets when using ME5-Base inversion models. Particularly noteworthy is the significant performance boost observed, especially evident when evaluating NQ-trained ME5-base model~\textsc{me\_nq} on MTG-DE, resulting in a remarkable 128.11\% performance gain.

It is interesting that multilingual LMs reconstruct texts in the language of training data, while monolingual language model (\textsc{GTR}) reconstruct texts mostly in the target language. 
This highlights the differences of monolingual and multilingual LMs, and warrants further research for future work.

\begin{table}[ht!]

\begin{subtable}[h]{0.5\textwidth}
\resizebox{\textwidth}{!}{
         \centering

     \begin{tabular}{c|ccc}
     \hline 
            & MTG-FR  & MTG-DE & MTG-ES \\\hline 
      \textbf{GTR-Base}  &  & & \\
      Base   & 4.39 (0.7581) & 3.22 (0.7052) & 4.74 (0.7134)    \\
    Vec2Text   & \textbf{10.91 (0.8833)} &  \textbf{6.46 (0.8138)} &  \textbf{10.84  (0.9020)}\\ 
    AdTrans  &  10.48  ($\downarrow$-3.92\%)  & 6.15 ($\downarrow$-4.84\%) & 9.95 ($\downarrow$-1.67\%) \\
    
    \hline 
    \textbf{ME5-Base}  &  & & \\
    Base &   3.13 (0.9513)  &  2.73 (0.9298) &  3.64 (0.9293)  \\
    Vec2Text &   6.46  (0.9487)   &  5.37  (0.9107) &  5.91  (0.8963) \\ 
    AdTrans &  \textbf{13.40  ($\uparrow$107.32\%)}  &  \textbf{ 8.54  ($\uparrow$59.21\%)}  & \textbf{ 11.87  ($\uparrow$100.79 \%)}  \\
    \hline 
    \end{tabular}
    
  }
  \caption{Cross-lingual cross-domain evaluation with monolingual models trained on NQ. }
    \label{tab:crosslingual_eval_nq2mtg}
    
    \end{subtable}
    
\hfill

\begin{subtable}[h]{0.5\textwidth}
\resizebox{\textwidth}{!}{
    \begin{tabular}{c|ccc}
    \hline 
      $\rightarrow$NQ  &  \textbf{\textsc{me5\_fr}} & \textbf{\textsc{me5\_de}}  & \textbf{\textsc{me5\_es}} \\\hline
       Base   & 2.60 (0.96) & 2.80 (0.8790) & 2.32 (0.9266)  \\
       Vec2Text& 4.00 (0.9441)  &  5.13 (0.9374)  &  3.41 (0.9380)  \\
       AdTrans &   \textbf{8.11 ($\uparrow$102.50\%)}  &   \textbf{10.18 ($\uparrow$98.49\%)}  &  \textbf{6.07 ($\uparrow$78.04\%)}  \\
       \hline
    \end{tabular}
    }
    \caption{Cross-lingual cross-domain evaluation on NQ with monolingual models trained on MTG datasets.}
    \label{tab:cross-domain_eval_mtg2nq}
\end{subtable}

    \caption{Cross-lingual evaluation using BLEU score and Cosine Similarity (in the brackets) for Base and Vec2Text models by correction steps of 50 with 8 sbeam. The BLEU scores and their growth (in the brackets) compared with BLEU scores on Vec2Text models are reported for AdTrans strategy for each model. $\uparrow$ indicates performance gain while the $\downarrow$ indicates performance loss. The result with the highest BLEU score with each evaluated model on each dataset  is in bold.}
    \label{tab:crosslingual_eval_all}
    
\end{table}

\section{Qualitative Analysis}\label{sec:qualitative_analysis}

\subsection{Multilingual Text Reconstruction}
We conduct qualitative analysis on multilingual text reconstruction using parallel samples. Table~\ref{tab:multi_es_fr} shows the French and Spanish samples, in comparison to Table~\ref{multi_en_de_qualitative}, samples in English and German.
The samples are evaluated on~\textsc{me5\_multi}. 
By Step 2, French sentence is already reconstructed with one word mismatch, however, the whole sentence is only fully reconstructed by correction step 50 + 4 sbeam. 
The cosine similarity is high from step 1, i.e., 0.9892, compared to the sample in English, i.e., 0.8056 and in German, i.e., 0.8704. 
While English and German samples are fully reconstructed by step 6 and 7.
As argued, the approximation approach with cosine similarity seems to be more effective for models rendering lower cosine similarity from initial steps.
However, from observations, \textsc{me5} models reconstructs closer embeddings across languages from the start.

\subsection{Cross-lingual Text Reconstruction}

We further conduct qualitative analysis on cross-lingual text reconstruction, aided by AdTrans.
As shown in Table~\ref{tab:crosslingual_qualitative}, the four way multilingual samples are used, all represent the same meaning. 
Each sample is evaluated by ME5-base inversion models trained on other three languages separately.

Consistent with previous quantitative analysis, the cross-lingual reconstruction is difficult, and the BLEU scores are consistently low. 
With AdTrans, the BLEU scores are overly boosted, with an exception of evaluating Spanish sample with ~\textsc{me5\_en}.
In this example, the highest performance gain is adding AdTrans for evaluating English sample with \textsc{me5\_de}.

The intention of adding AdTrans is to improve the utility of current string-matching metrics in cross-lingual attack setting, while also expose the inadequacy of such metrics in terms of LLMSec. With this example, there is essential information leakage in each evaluation that can not be captured even after applying AdTrans.

\begin{table*}[ht]
\centering
  \resizebox{0.9\textwidth}{!}{
    \begin{tabular}{llrr}
    \hline 
\textbf{Step}  & \textbf{Text} & \textbf{BLEU} & \textbf{COS} \\\hline 
Input &	\textbf{ford doit rappeler 1,3 million de suv en raison des gaz d'échappement} & & \\	
Step 1 &	ford doit rappeler 1,3 million de suv en raison \colorbox{pink}{du} gaz \colorbox{pink}{d'absorption}	 &68.12 &	\underline{0.9892}\\
Step 2	 &ford doit rappeler 1,3 million de suv en raison \colorbox{pink}{du} gaz d'échappement &	76.12 &	0.9712\\
Step 3	 &ford doit rappeler 1,3 million de suv en raison \colorbox{pink}{du} gaz d'échappement	 &76.12	 &0.9992\\
Step 4 &	ford doit rappeler 1,3 million de suv en raison \colorbox{pink}{du} gaz d'échappement	 &76.12 &	0.9712\\
Step 5	 &ford doit rappeler 1,3 million de suv en raison \colorbox{pink}{du} gaz d'échappement &	76.12	 &0.9992\\
Step 6 &	ford doit rappeler 1,3 million de suv en raison \colorbox{pink}{du} gaz d'échappement	 &76.12 &	0.9712\\
Step 7 &	ford doit rappeler 1,3 million de suv en raison \colorbox{pink}{du} gaz d'échappement	 &76.12 &	0.9712\\
Step 50  &	ford doit rappeler 1,3 million de suv en raison \colorbox{pink}{du} gaz d'échappement&	76.12 &0.9992\\
Step 50 + 4 sbeam &	\colorbox{green}{ford doit rappeler 1,3 million de suv en raison des gaz d'échappement}	 &\textbf{100} &	\textbf{0.9915}\\

\hline

Input &	\textbf{ford instó a retirar 1.3 millones suvs por el escape de humos} &   &  \\
Step 1 &ford \colorbox{pink}{imploró} \colorbox{yellow}{el} 1,3 millones \colorbox{pink}{de} suvs \colorbox{pink}{en la salida} de humos & 8.91&	\underline{0.9491} \\

Step 2 & ford \colorbox{pink}{advirtió}  \colorbox{yellow}{el} 1,3 millones \colorbox{yellow}{de humos} \colorbox{pink}{selevados de suvs al elimin} &	8.91 &	0.8213 \\

Step 3 & ford \colorbox{pink}{se advirtió por eliminar} 1,3 millones de humos a suvs a sale	& 8.45 & 0.9634 \\
Step 4 & ford \colorbox{pink}{se advirtió por el rescate de} 1,3 millones \colorbox{pink}{de} suvs por \colorbox{pink}{hum}	& 9.67&	0.9552\\

Step 5 & ford \colorbox{pink}{se advirtió que} 1,3 millones \colorbox{pink}{de} suvs \colorbox{pink}{se escaparon} por \colorbox{pink}{humo} &5.06 &0.9696 \\

Step 6 & ford \colorbox{pink}{se instó a la sépara de} 1,3 millones \colorbox{pink}{de} suvs por humos & 10.39 &	0.9045 \\

Step 7 & ford \colorbox{pink}{instó a los} 1,3 millones \colorbox{pink}{de} suvs \colorbox{pink}{a salir del humo revapor} &	13.67 & 0.9481\\
Step 50 &ford \colorbox{pink}{instó a la salida de} 1.3 millones \colorbox{pink}{de} suvs por el \colorbox{pink}{humo}  &	22.63 &	0.9794\\
Step 50 + 4 sbeam  & ford \colorbox{pink}{instó a la salida de} 1.3 millones \colorbox{pink}{de} suvs \colorbox{pink}{con} humos \colorbox{pink}{para elimin} &14.95 &	0.831\\
Step 50 + 8 sbeam & \colorbox{green}{ford instó a retirar 1.3 millones suvs por el escape de humos}& \textbf{100} & \textbf{1.0000}\\

\hline

    \end{tabular}
    }
    \caption{Qualitative Analysis of Reconstructing Multilingual Parallel Texts in French and Spanish using \textsc{me5\_multi}. \textbf{Step} are the correction steps from Step 1 (initial hypothesis) to Step 50 + 4/8 sbeam for the correct inversions. The colored boxes indicate \colorbox{yellow}{misplaced tokens}, \colorbox{pink}{wrong tokens}, and \colorbox{green}{exact matches}. The best results for metrics are in \textbf{bold}. Initial cosine similarity is \underline{underlined}.} 
    \label{tab:multi_es_fr}
\end{table*}

\begin{table*}[!h]
    \centering
  \resizebox{\textwidth}{!}{
    \begin{tabular}{l|p{7cm}rrp{7cm}l}
    \hline
        \textbf{Model} & \textbf{Text} & \textbf{BLEU} & \textbf{COS} & \textbf{AdTrans} &\textbf{BLEU} \\\hline

Input &\multicolumn{5}{l}{\textbf{ford urged to recall 1.3 million suvs over exhaust fumes.}} \\ 
\textsc{me5\_es} &	ford insiste on-reclame a 1,3 millones de suvs.	&5.02	&0.8922	&\colorbox{green}{ford} \colorbox{cyan}{insists on-reclaiming} me to \colorbox{green}{1.3 million suvs.}&	16.59$\uparrow$\\

\textsc{me5\_fr} &	ford exhorte recall of blow 'parmi les 1,3 million de suvs.	&5.06	&0.9717& \colorbox{green}{ford} \colorbox{cyan}{urges} \colorbox{green}{recall} of blow 'among the \colorbox{green}{1.3 million suvs.}&	16.59$\uparrow$ \\

\textsc{me5\_de} & ford appelliert an recall of 1,3 millionen suvs über fume. &	5.3	&0.8866 &	\colorbox{green}{ford} \colorbox{cyan}{appeals} \colorbox{green}{to recall} of \colorbox{green}{1.3 million suvs over} fume \colorbox{green}{.}	&29.98 $\uparrow$ \\

\hline

Input & \multicolumn{5}{l}{\textbf{ford doit rappeler 1,3 million de suv en raison des gaz d'échappement.}}			\\
\textsc{me5\_en}	& ford notices that 1.3 million suvs get recalled for gas-shock.&	4.11	&0.9276&	 \colorbox{green}{ford} remarque que \colorbox{green}{1,3 million de} \colorbox{cyan}{suvs} sont \colorbox{cyan}{rappelés} pour le choc au \colorbox{green}{gaz} \colorbox{green}{.} &	11.72$\uparrow$\\

\textsc{me5\_es}	& ford se debe a recordar 1,3 millones de suv por el evacuación de gas.	&6.61	&0.903&	 \colorbox{green}{ford} est dû à la mémoire de  \colorbox{green}{1,3 million de suv }pour l'évacuation du gaz.&	17.66$\uparrow$\\

\textsc{me5\_de} &	ford cite 1,3 millionen gas suv, weshalb sie die abmeldung verpassen sollten.&	4.02&	0.903&	 \colorbox{green}{ford} cite  \colorbox{green}{1,3 millions} de  \colorbox{green}{gaz}  \colorbox{green}{suv}, c'est \colorbox{cyan}{pourquoi} vous devriez rater l'annulation.&	4.15$\uparrow$\\
\hline

Input &\multicolumn{5}{l}{\textbf{ford wird aufgefordert 1,3 millionen suvs wegen abgasen zurückzurufen.}} \\ 
					
\textsc{me5\_en} &	ford has demanded that ford call back 1.3 million agressive suvs.&	4.46&	0.9049	& \colorbox{green}{ford} hat \colorbox{cyan}{gefordert}, dass ford\colorbox{green}{1,3 millionen}agressive\colorbox{green}{suvs}\colorbox{cyan}{zurückruft}\colorbox{green}{.} &	8.64$\uparrow$ \\

\textsc{me5\_es} &	ford ha exigido un apagón de 1.3 millones de suvs por regreso.	&4.07&	0.891&	\colorbox{green}{ford}hat einen stromausfall von\colorbox{green}{1,3 millionen suvs} \colorbox{cyan}{auf dem rückweg gefordert}\colorbox{green}{.} &	13.15$\uparrow$ \\
\textsc{me5\_fr} &	ford réclame une récharge de 1,3 million de suvs en raison des agressions. &	4.02 &	0.889	&	\colorbox{green}{ford} \colorbox{cyan}{fordert} eine aufladung von \colorbox{green}{1,3 millionen suvs} wegen der übergriffe\colorbox{green}{.} &	23.80 $\uparrow$\\
\hline

Input & 	\multicolumn{5}{l}{\textbf{ford instó a retirar 1.3 millones suvs por el escape de humos.}}\\

\textsc{me5\_en} &	ford vows to save 1.3 million suvs of smoke ford was expelled. &		4.37 &		0.8476 &		vado votos para salvar \colorbox{green}{1,3 millones} de \colorbox{green}{suvs} de humo vado fue expulsado.	 &	4.05 $\downarrow$\\

\textsc{me5\_fr} & 	ford a revendiqué 1 milliard de smaux de fumée pour le sortir de ses suvs.	&3.66&	0.9183&	\colorbox{green}{ford} reivindicó mil millones de smalls de humo para sacarlo de sus súbditos.&	4.05 $\uparrow$\\

\textsc{me5\_de} &	ford appellierte die befreiung mit dem rauch es gibt 1,3 milliarden suvs.&	4.07	&0.8642	& \colorbox{green}{ford} apeló a la liberación con el humo hay \colorbox{green}{1,3} mil millones de \colorbox{green}{suvs}.	&4.31$\uparrow$\\\hline

    \hline 
    \end{tabular}}

    \caption{Qualitative Analysis of Cross-lingual Text Reconstruction using monolingual ME5-base models. \textbf{Text} shows the input and the reconstructed texts by Step 50 + 8 sbeam in the regarding languages, and subsequent the metrics for evaluation (\textbf{BLEU} and \textbf{COS}). \textbf{AdTrans} shows the translation of reconstructed text into the target language. The second \textbf{BLEU} evaluates the translated text to the original with $\uparrow$ indicating performance gains. The colored boxes indicate \colorbox{green}{matched tokens} and \colorbox{cyan}{information leakages}.}
    \label{tab:crosslingual_qualitative}
    
\end{table*}

\section{Full Defense Results}~\label{app:full_defense}
\subsection{Noise Insertion Defense}~\label{app:gaussian_noise}
Following~\citep{morris-etal-2023-text}, the noisy embedding model is defined as following:
\begin{equation}\label{eq:gaussian}
    \phi_{noisy}(x)= \phi(x)+\lambda \cdot \epsilon, \epsilon\in \mathbb{N}(0,1)
\end{equation}
where $\lambda$ is a hyperparameter controlling the amount of noise injected.

\subsection{Language Neutrality of Inversion Models }

Drawing inspiration from \citet{libovicky-etal-2020-language}, we delve into the impact of \textit{language-agnostic} embeddings on retrieval and reconstruction performance. This is achieved by isolating the \textit{language-specific} component, represented by the mean of the embeddings, which serves to identify the language of the representations. Conversely, we extract the \textit{language-agnostic} component by subtracting the mean embeddings, thereby capturing the essence of the text in a language-independent manner. 

We present the performance of language-agnostic component on GTR-base and ME5-base models across BEIR and CLIRMatrix benchmarks in Table~\ref{tab:beir_retrieval_performance}, \ref{tab:beir_reconstruction},~\ref{tab:beir_combo}, and~\ref{tab:clirmatrix_retrieval},~\ref{tab:clirmatrix_reconstruction},~\ref{tab:clirmatrix_combo}.

Consistently, our findings demonstrate that language-agnostic embeddings either outperform or perform equally well compared to the original embeddings in retrieval tasks. However, while there is only a slight degradation in performance for text reconstruction on the CLIRMatrix benchmark and with ME5-base models on the BEIR benchmark, the reconstruction performance experiences a notable 20\% decline with the GTR-base model on the BEIR benchmark. This indicate that the distinction of language-specific and language-agnostic component is more salient for multilingual models.

\subsection{Results on BEIR Benchmark}
We reproduce the retrieval and reconstruction on GTR-base models across 12 BEIR tasks from~\citep{morris-etal-2023-text}, excluding the four private datasets. 
Moreover, we implement retrieval on ME5-base models. 
The full defense results for retrieval performance and reconstruction tasks are shown in Table~\ref{tab:beir_retrieval_performance}, \ref{tab:beir_reconstruction} and \ref{tab:beir_combo}.

\subsection{Results on CLIRMatrix Benchmark}
To evaluate the cross-lingual scenario in retrieval and reconstruction on monolingual and multilingual models, we implement cross-lingual retrieval and text reconstruction across 12 cross-lingual datasets constructed from MULTI-8 of CLIRMatrix~\citep{sun-duh-2020-clirmatrix}. 

Let $q$ be a query in language $L_{query}$ and $d$ be a document in language $L_{doc}$. 
In our scenario, the cross-lingual retrieval task involves retrieving the document in language $L_{doc}$ when presented with a query in language $L_{query}$ within the nearest neighbor retrieval framework. 
For our evaluation, the cross-lingual datasets are constructed with the triple $(q^{L_{query}}, d^{L_{doc}})$, where $L_{query}\in \{en, fr, de, es\}$ and $L_{doc}\in \{en, fr, de, es\}$, and $L_{query}\neq L_{doc}$.
We implement retrieval and reconstruction both on GTR-base and ME5-base models.

The full defense results for retrieval performance and reconstruction tasks are shown in Table~\ref{tab:clirmatrix_retrieval}, \ref{tab:clirmatrix_reconstruction} and \ref{tab:clirmatrix_combo}.

\begin{table*}[ht]
    \centering
    \resizebox{\textwidth}{!}{
    \begin{tabular}{l|cccccccccccc}
    \hline
        & arguana & climate-fever & dbpedia-entity & fiqa & msmarco & nfcorpus & nq & quora & scidocs & scifact & trec-covid & webis-touche2020 \\ \hline
        
        \textbf{GTR} & ~ & ~ & ~ & ~ & ~ & ~ & ~ & ~ & ~ & ~ & ~ & ~ \\ 
         $\lambda$& ~ & ~ & ~ & ~ & ~ & ~ & ~ & ~ & ~ & ~ & ~ & ~ \\ 
        0 & 0.3278 & 0.1355 & 0.3058 & 0.2080 & 0.6466 & 0.2392 & 0.3060 & 0.8794 & 0.0951 & 0.2472 & 0.3757 & 0.2335 \\
       0.001 & 0.3276 & 0.1358 & 0.3079 & 0.2089 & 0.6480 & 0.2392 & 0.3056 & 0.8791 & 0.0948 & 0.2481 & 0.3775 & 0.2309 \\ 

        0.01 & 0.3203 & 0.1307 & 0.2993 & 0.2044 & 0.6328 & 0.2352 & 0.2993 & 0.8747 & 0.0930 & 0.2417 & 0.3702 & 0.2314 \\ 
        0.1 & 0.0059 & 0.0000 & 0.0003 & 0.0008 & 0.0026 & 0.0147 & 0.0001 & 0.0041 & 0.0011 & 0.0011 & 0.0049 & 0.0000 \\ 
        1 & 0.0008 & 0.0000 & 0.0000 & 0.0000 & 0.0000 & 0.0081 & 0.0000 & 0.0000 & 0.0003 & 0.0000 & 0.0000 & 0.0000 \\\hdashline
         Masking & 0.32724 & 0.13585 & 0.3057 & 0.20788 & 0.6463 & 0.23954 & 0.30574 & 0.87937 & 0.09549 & 0.2457 & 0.37763 & 0.23341 \\ 
        Lang-agnostic & 0.3275 & 0.13502 & 0.30589 & 0.20787 & 0.64664 & 0.23913 & 0.30564 & 0.87929 & 0.09542 & 0.24838 & 0.37687 & 0.23212  \\ \hline

         \textbf{ME5} & ~ & ~ & ~ & ~ & ~ & ~ & ~ & ~ & ~ & ~ & ~ & ~ \\
          $\lambda$& ~ & ~ & ~ & ~ & ~ & ~ & ~ & ~ & ~ & ~ & ~ & ~ \\ 
        0 & 0.3002 & 0.1441 & 0.3389 & 0.2155 & 0.6446 & 0.2509 & 0.3344 & 0.8788 & 0.1180 & 0.2876 & 0.4836 & 0.2208 \\ 
         0.001 & 0.3014 & 0.1433 & 0.3368 & 0.2155 & 0.6449 & 0.2506 & 0.3351 & 0.8783 & 0.1174 & 0.2871 & 0.4818 & 0.2241 \\ 
        0.01 & 0.2725 & 0.1267 & 0.3094 & 0.1936 & 0.6257 & 0.2368 & 0.3089 & 0.8634 & 0.1055 & 0.2509 & 0.4363 & 0.2141 \\ 
       
        0.1 & 0.0006 & 0.0000 & 0.0001 & 0.0004 & 0.0000 & 0.0098 & 0.0000 & 0.0002 & 0.0006 & 0.0010 & 0.0000 & 0.0000 \\
        1 & 0.0005 & 0.0000 & 0.0000 & 0.0000 & 0.0000 & 0.0108 & 0.0000 & 0.0000 & 0.0003 & 0.0010 & 0.0000 & 0.0000 \\ \hdashline
        Masking & 0.30038 & 0.14403 & 0.33753 & 0.21603 & 0.64487 & 0.2512 & 0.33473 & 0.87858 & 0.11747 & 0.28666 & 0.4837 & 0.22062 \\ 
        Lang-agnostic & 0.30021 & 0.14411 & 0.33891 & 0.2155 & 0.64459 & 0.25092 & 0.33442 & 0.87877 & 0.11793 & 0.28755 & 0.48357 & 0.22076 \\ \hline

    \end{tabular}}
    \caption{BEIR performance (NDCG@10) for GTR-base and ME5-base at varying level of random noise (32 tokens).}
    \label{tab:beir_retrieval_performance}
\end{table*}

\begin{table*}[!h]
    \centering
     \resizebox{\textwidth}{!}{
    \begin{tabular}{l|cccccccccccc}
    \hline
    \toprule
         & arguana & climate-fever & dbpedia-entity & fiqa & msmarco & nfcorpus & nq & quora & scidocs & scifact & trec-covid & webis-touche2020 \\ \hline
   \textbf{GTR} &  & ~ & ~ & ~ & ~ & ~ & ~ & ~ & ~ & ~ & ~ & ~ \\ 
   $\lambda$ & & ~ & ~ & ~ & ~ & ~ & ~ & ~ & ~ & ~ & ~ & ~ \\ 
   0 & 60.43 & 82.65 & 68.26 & 41.12 & 61.72 & 67.52 & 80.98 & 43.87 & 63.6 & 65.64 & 65.4 & 37.76 \\ 
        0.001 & 47.23 & 72.73 & 53.93 & 33.27 & 49.08 & 53.22 & 65.18 & 42.5 & 48.92 & 53.36 & 53.31 & 30.88 \\
        0.01 & 7.59 & 16.26 & 11.6 & 7.13 & 9.85 & 8.26 & 10.52 & 15.3 & 6.86 & 8.1 & 8.91 & 8.51 \\ 
        0.1 & 1.71 & 1.92 & 1.83 & 1.65 & 1.76 & 1.74 & 1.77 & 1.64 & 1.71 & 1.78 & 1.72 & 1.72 \\
        1 & 1.48 & 1.58 & 1.51 & 1.41 & 1.63 & 1.51 & 1.53 & 0.98 & 1.49 & 1.59 & 1.5 & 1.4 \\ \hdashline
        Masking & 3.69 & 7.71 & 4.52 & 3.68 & 4.37 & 3.89 & 4.42 & 9.36 & 3.06 & 3.38 & 3.63 & 4.44 \\ 
        Lang-agnostic & 49.15 & 70.96 & 58.22 & 32.37 & 47.69 & 53.04 & 67.29 & 40.39 & 52.53 & 52.1 & 54.56 & 31.9 \\ 
        
        \midrule

           \textbf{ME5\_NQ} &  & ~ & ~ & ~ & ~ & ~ & ~ & ~ & ~ & ~ & ~ & ~ \\ 
                $\lambda$&  & ~ & ~ & ~ & ~ & ~ & ~ & ~ & ~ & ~ & ~ & ~ \\ 
    0 & 46.75 & 63.29 & 63.21 & 30.57 & 51.24 & 54.35 & 71.49 & 24.85 & 51.18 & 52.76 & 50.8 & 28.44 \\
        0.001 & 44.62 & 35.28 & 39.01 & 30.94 & 42.67 & 54.09 & 45.31 & 17.56 & 52.15 & 53.04 & 51.03 & 30.96 \\ 
        0.01 & 35.8 & 30.33 & 34 & 25.63 & 33.3 & 45.52 & 38.34 & 15.95 & 40.86 & 43.61 & 40.88 & 24.69 \\ 
        0.1 & 3.8 & 5.11 & 4.84 & 3.4 & 4.27 & 4 & 4.63 & 3 & 3.58 & 3.64 & 3.65 & 3.34 \\ 
        1 & 1.94 & 2.11 & 1.92 & 1.82 & 2.04 & 2.05 & 2.12 & 1.2 & 1.9 & 1.95 & 1.98 & 1.89 \\ \hdashline
        Masking & 9.68 & 12.72 & 13.06 & 8.85 & 11.09 & 11.18 & 11.98 & 10.89 & 9.06 & 9.39 & 9.57 & 8.97 \\ 
        Lang-agnostic & 43.41 & 35.12 & 38.49 & 30.27 & 39.76 & 54.64 & 45.29 & 17.92 & 50.94 & 51.56 & 48.8 & 28.23 \\  \hline

         \textbf{\textsc{me5\_en}} &  & ~ & ~ & ~ & ~ & ~ & ~ & ~ & ~ & ~ & ~ & ~ \\ 
                $\lambda$&  & ~ & ~ & ~ & ~ & ~ & ~ & ~ & ~ & ~ & ~ & ~ \\ 
         0 & 39.29 & 54.51 & 32.24 & 32.68 & 39.76 & 37.9 & 55.09 & 76.92 & 33.62 & 28.5 & 32.87 & 37.04 \\
        0.001 & 38.36 & 53.84 & 31.71 & 32.34 & 38.17 & 37.34 & 54.76 & 77.05 & 33.24 & 28.48 & 31.98 & 37.34 \\ 
        0.01 & 33.01 & 43.22 & 28.43 & 28.24 & 33.89 & 33.11 & 46.93 & 65.83 & 28.94 & 24.86 & 27.98 & 31.28 \\ 
        0.1 & 4.31 & 5.79 & 5.26 & 3.63 & 4.7 & 4.43 & 5.5 & 4.95 & 3.58 & 3.75 & 4.01 & 4.19 \\ 
        1 & 1.79 & 1.95 & 1.83 & 1.71 & 1.85 & 1.93 & 1.99 & 1.23 & 1.78 & 1.8 & 1.78 & 1.65 \\ \hdashline
        Masking & 10.98 & 13.55 & 11.87 & 10.11 & 11.79 & 10.61 & 15.48 & 17.65 & 8.24 & 8.14 & 9.51 & 10.16 \\ 
        Lang-agnostic & 38.77 & 51.98 & 30.87 & 31.9 & 37.61 & 36.49 & 52.67 & 74.27 & 31.67 & 28.65 & 30.18 & 36.1 \\\hline
        
        \textbf{\textsc{me5\_MULTI}} &  & ~ & ~ & ~ & ~ & ~ & ~ & ~ & ~ & ~ & ~ & ~ \\ 
        $\lambda$&  & ~ & ~ & ~ & ~ & ~ & ~ & ~ & ~ & ~ & ~ & ~ \\ 
      0 & 23.02 & 31.38 & 21.89 & 20.55 & 25.39 & 22.45 & 35.55 & 62.65 & 18.99 & 16.71 & 19.89 & 23.28 \\ 
        0.001 & 23.54 & 31.61 & 22.46 & 20.05 & 25.04 & 22.58 & 35.38 & 62.24 & 19.02 & 16.95 & 18.76 & 22.72 \\
        0.01 & 20.2 & 26.36 & 20.06 & 16.7 & 21.59 & 19.69 & 30.49 & 52.94 & 15.93 & 14.95 & 17.65 & 20.12 \\ 
        0.1 & 3.62 & 4.66 & 4.4 & 3.31 & 4.05 & 3.84 & 4.22 & 4.21 & 3.08 & 3.5 & 3.62 & 3.6 \\ 
        1 & 0.94 & 1 & 1.23 & 0.92 & 1.05 & 1.01 & 1.18 & 0.61 & 0.98 & 0.97 & 0.99 & 0.92 \\ \hdashline
        Masking & 7.76 & 9.7 & 8.98 & 7.19 & 8.85 & 7.26 & 10.48 & 14.33 & 6.18 & 6.22 & 6.65 & 7.17 \\ 
        Lang-agnostic & 22.83 & 31.08 & 22.09 & 19.13 & 24.07 & 21.52 & 33.77 & 60.15 & 18.13 & 16.79 & 18.66 & 22.95 \\

        \bottomrule
        
    \end{tabular}}
        \caption{BEIR Text Reconstruction performance (BLEU score) for monolingual and multilingual inversion models at varying level of random noise (32 tokens).}
        \label{tab:beir_reconstruction}

\end{table*}

\begin{table*}[!h]
\centering
\resizebox{0.7\textwidth}{!}{
    \begin{tabular}{l|c|c||c|ccc}

\toprule
      &\multicolumn{2}{c}{\textbf{GTR-Based}} & \multicolumn{4}{c}{\textbf{ME5-Based}}  \\
    
       \textbf{Defenses} & IR (NDCG@10)  & \textsc{gtr} & IR (NDCG@10) & \textsc{me5\_nq} & \textsc{me5\_en} & \textsc{me5\_multi}  \\ \hline
        \midrule
        $\lambda$ & & & & & & \\
        0 & 0.3333 & 61.58 & 0.3514 & 49.08 & 41.7 & 26.81 \\ 
        0.001 & 0.3336 & 50.3 & 0.3514 & 41.39 & 41.22 & 26.7 \\ 
        0.01 & 0.3277 & 9.91 & 0.3286 & 34.08 & 35.48 & 23.06 \\
        0.1 & 0.003 & 1.75 & 0.0011 & 3.94 & 4.51 & 3.84 \\ 
        1 & 0.0008 & 1.47 & 0.001 & 1.91 & 1.77 & 0.98 \\ \hdashline
        Masking & 0.3333 & 4.68 & 0.3513 & 10.54 & 11.51 & 8.4 \\ 
        Lang-agnostic & 0.3333 & 50.85 & 0.3514 & 40.37 & 40.1 & 25.93 \\ 
        \bottomrule
    \end{tabular}}
    \caption{BEIR Retrieval Performance  (NDCG@10) and Reconstruction performance (BLEU) (mean across tasks) with GTR-based (left) and  ME5-based (right) models across varying level of random noises and defense algorithms.}
\label{tab:beir_combo}
\end{table*}


 \begin{table*}[ht]
    \centering
       \resizebox{\textwidth}{!}{
    \begin{tabular}{l|rrr|rrr|rrr|rrr}
\hline
  \toprule
   $L_{query}$ &\multicolumn{3}{c}{\textbf{English}} & \multicolumn{3}{c}{\textbf{French}} & \multicolumn{3}{c}{\textbf{German}} &  \multicolumn{3}{c}{\textbf{Spanish}} \\
    
  $L_{doc}$ & \textsc{French} & \textsc{German}  & \textsc{Spanish} & \textsc{English} & \textsc{German} & \textsc{Spanish} & \textsc{English} & \textsc{French} & \textsc{Spanish} & \textsc{English} & \textsc{French} & \textsc{German} \\
  
  \midrule

         \textbf{GTR}& ~ & ~ & ~ & ~ & ~ & ~ & ~ & ~ & ~ & ~ & ~ & ~ \\ 
        
        $\mathbf{\lambda}$  & ~ & ~ & ~ & ~ & ~ & ~ & ~ & ~ & ~ & ~ & ~ & ~ \\ 
        0 & 0.19407 & 0.26324 & 0.24222 & 0.13205 & 0.14329 & 0.13589 & 0.1243 & 0.08702 & 0.1177 & 0.10308 & 0.088 & 0.10494 \\
        0.001 & 0.19377 & 0.2633 & 0.24108 & 0.13237 & 0.1435 & 0.13627 & 0.12476 & 0.08786 & 0.11741 & 0.10301 & 0.08805 & 0.1055 \\ 
        0.01 & 0.18651 & 0.25203 & 0.2326 & 0.12676 & 0.13617 & 0.13298 & 0.11846 & 0.07997 & 0.11234 & 0.09713 & 0.08234 & 0.09794 \\ 
        0.1 & 0 & 0 & 0 & 0 & 4.00E-05 & 0 & 0 & 0 & 0 & 0.00016 & 0 & 0.00023 \\ 
        1 & 0 & 0 & 0 & 0 & 0 & 0 & 0 & 0 & 0 & 0 & 0 & 0 \\ \hdashline
        Masking & 0.19433 & 0.26272 & 0.24193 & 0.13237 & 0.14322 & 0.13579 & 0.12402 & 0.0871 & 0.11791 & 0.10337 & 0.08818 & 0.10477 \\ 
        Lang-agnostic & 0.19439 & 0.26265 & 0.24206 & 0.13217 & 0.14284 & 0.13636 & 0.1241 & 0.08735 & 0.11766 & 0.10313 & 0.08845 & 0.10528 \\ \hline

    \textbf{ME5\_MULTI}& ~ & ~ & ~ & ~ & ~ & ~ & ~ & ~ & ~ & ~ & ~ & ~ \\ 
        
        $\mathbf{\lambda}$  & ~ & ~ & ~ & ~ & ~ & ~ & ~ & ~ & ~ & ~ & ~ & ~ \\ 
        0 & 0.2861 & 0.3739 & 0.4141 & 0.2932 & 0.2598 & 0.3121 & 0.2878 & 0.1940 & 0.2875 & 0.2860 & 0.2181 & 0.2425 \\ 
        0.001 & 0.2853 & 0.3740 & 0.4124 & 0.2935 & 0.2603 & 0.3121 & 0.2876 & 0.1931 & 0.2873 & 0.2850 & 0.2181 & 0.2433 \\ 
        0.01 & 0.2484 & 0.3374 & 0.3731 & 0.2580 & 0.2271 & 0.2779 & 0.2583 & 0.1654 & 0.2590 & 0.2452 & 0.1811 & 0.2121 \\ 
        0.1 & 0 & 0 & 0.0001 & 0 & 0 & 0.0002 & 0 & 0 & 0 & 0 & 0 & 0 \\ 
        1 & 0 & 0 & 0 & 0 & 0 & 0.0002 & 0 & 0 & 0.0001 & 0 &  & 0 \\ \hdashline
          Masking  & 0.2861 & 0.3755 & 0.4129 & 0.2933 & 0.2594 & 0.3125 & 0.2882 & 0.1935 & 0.2877 & 0.2862 & 0.2179 & 0.2426 \\ 
        
        Lang-agnostic & 0.2859 & 0.3740 & 0.4142 & 0.2933 & 0.2598 & 0.3125 & 0.2878 & 0.1939 & 0.2874 & 0.2862 & 0.2180 & 0.2425 \\ 
        
        \bottomrule
    \end{tabular}}

    \caption{CLIRMatrix (multi8) performance (NDCG@10) for GTR-base and ME5-base at varying defense mechanisms (32 tokens).}
    \label{tab:clirmatrix_retrieval}
    
\end{table*}

\begin{table*}[!h]
    \centering
       \resizebox{\textwidth}{!}{
    \begin{tabular}{l|rrr|rrr|rrr|rrr}
\hline
  \toprule
   $L_{query}$ &\multicolumn{3}{c}{\textbf{English}} & \multicolumn{3}{c}{\textbf{French}} & \multicolumn{3}{c}{\textbf{German}} &  \multicolumn{3}{c}{\textbf{Spanish}} \\
    
 $L_{doc}$  & \textsc{French} & \textsc{German}  & \textsc{Spanish} & \textsc{English} & \textsc{German} & \textsc{Spanish} & \textsc{English} & \textsc{French} & \textsc{Spanish} & \textsc{English} & \textsc{French} & \textsc{German} \\
  
  \midrule

 \textbf{GTR} & ~ & ~ & ~ & ~ & ~ & ~ & ~ & ~ & ~ & ~ & ~ & ~ \\ 
        $\mathbf{\lambda}$  & ~ & ~ & ~ & ~ & ~ & ~ & ~ & ~ & ~ & ~ & ~ & ~ \\
        0 & 10.78 & 10.99 & 12.2 & \textbf{30.97} & 12.9 & 14.57 & \textbf{28.55} & 10.91 & 11.58 & \textbf{29.34} & 9.85 & 8.77  \\ 
        0.001 & 10.74 & 10 & 11.87 & 25.32 & 11.19 & 13.22 & 24.32 & 10.34 & 10.75 & 24.82 & 9.22 & 8.97  \\ 
        0.01 & 5.53 & 5.88 & 5.95 & 7.74 & 5.53 & 6.12 & 6.24 & 4.56 & 5.07 & 7.55 & 4.95 & 4.78  \\ 
        0.1 & 1.34 & 1.42 & 1.25 & 1.23 & 0.8 & 0.67 & 0.98 & 0.74 & 0.58 & 1.02 & 0.63 & 0.59  \\ 
        1 & 0.58 & 0.49 & 0.32 & 0.77 & 0.35 & 0.26 & 0.78 & 0.55 & 0.34 & 0.73 & 0.41 & 0.41  \\ \hdashline
        Masking & 4.19 & 4.18 & 4.28 & 3.84 & 2.86 & 3.01 & 3.24 & 2.37 & 2.47 & 3.61 & 2.79 & 2.66 \\ 
        Lang-agnostic & 11.01 & 10.95 & 11.88 & 26.19 & 12.13 & 13.29 & 24.24 & 10.42 & 11.08 & 25.02 & 10.6 & 9.61  \\ \hline

    \textbf{ME5}& ~ & ~ & ~ & ~ & ~ & ~ & ~ & ~ & ~ & ~ & ~ & ~ \\ 
     $\mathbf{\lambda}$  & ~ & ~ & ~ & ~ & ~ & ~ & ~ & ~ & ~ & ~ & ~ & ~ \\
      0 & 11.86 & 11.1 & 17.87 & 15.11 & 12.79 & 17.68 & 14.5 & 13.66 & 17.44 & 14.34 & 13.67 & 11.99 \\ 
        0.001 & 12.63 & 10.49 & 17.12 & 15.43 & 12.63 & 17.19 & 14.39 & 13.56 & 17.21 & 14.79 & 14.4 & 11.84  \\ 
        0.01 & 11.27 & 9.78 & 14.9 & 13.48 & 10.99 & 15.06 & 13.65 & 12.1 & 15.38 & 14.07 & 12.3 & 11.61  \\ 
        0.1 & 2.27 & 2.23 & 2.56 & 2.97 & 2.4 & 2.78 & 2.7 & 2.1 & 2.47 & 2.92 & 2.31 & 2.48  \\ 
        1 & 0.53 & 0.44 & 0.5 & 0.68 & 0.57 & 0.5 & 0.62 & 0.49 & 0.46 & 0.66 & 0.55 & 0.47  \\ \hdashline
        Masking & 3.91 & 4.38 & 6.19 & 5.18 & 4.57 & 6.35 & 5.13 & 3.97 & 5.85 & 5.86 & 4.97 & 5.05 \\ 
        Lang-agnostic & 12.61 & 10.98 & 17.01 & 14.27 & 11.77 & 16.05 & 13.92 & 13.3 & 17.19 & 14.57 & 13.79 & 12.09 \\ 
        \bottomrule
    \end{tabular}}

    \caption{CLIRMatrix (multi8) Text Reconstruction Performance (BLEU score) for \textsc{GTR} and \textsc{me5\_multi} at varying defense mechanisms (32 tokens). The performances for \textsc{gtr} without noise on English doc are in \textbf{bold}, which boost the GTR's overall performance.}
    \label{tab:clirmatrix_reconstruction}
    
\end{table*}

\begin{table*}[ht]
    \centering
    \resizebox{0.6\textwidth}{!}{
    \begin{tabular}{l|cc||cc}

\toprule
      & \multicolumn{2}{c}{\textbf{GTR-Based}} & \multicolumn{2}{c}{\textbf{ME5-Based}}  \\
    
       \textbf{Defenses} & IR (NDCG@10) & \textsc{gtr} & IR (NDCG@10) & \textsc{me5\_multi}  \\ 
        \midrule
        $\lambda$ &  &  & & \\
        0 & 0.1447 & 15.95 & 0.2879 & 14.33 \\ 
        0.001 & 0.1447 & 14.23 & 0.2877 & 14.31 \\ 
        0.01 & 0.1379 & 5.83 & 0.2536 & 12.88 \\ 
        0.1 & 0.0000 & 0.94 & 0.0000 & 2.52 \\ 
        1 & 0.0000 & 0.50 & 0.0000 & 0.54 \\ \hdashline
        Masking & 0.1446 & 3.29 & 0.2880 & 5.12 \\ 
        Lang-agnostic & 0.1447 & 14.70 & 0.2879 & 13.96 \\ \hline 
    \end{tabular}
}
\caption{CLIRMatrix Retrieval Performance  (NDCG@10) and Reconstruction performance (BLEU) (mean across tasks) with GTR-based (left) and  ME5-based (right) models across varying level of random noises and defense algorithms.}
\label{tab:clirmatrix_combo}
\end{table*}

\end{document}